# VERSA – Video Event Recognition for Surveillance Applications

A Thesis Presented to the

Department of Computer Science

And the Faculty of the Graduate College

University of Nebraska

In Partial Fulfillment of the Requirements for the Degree

Masters of Science

University of Nebraska at Omaha

By

Stephen O'Hara

November 3rd, 2008

Supervisory Committee

Qiuming Zhu, Chair

William Mahoney

Prithviraj Dasgupta

Ken Dick

# VERSA – Video Event Recognition for Surveillance Applications

Stephen O'Hara, MS

University of Nebraska, 2008

Advisor: Qiuming Zhu

**Abstract**:

VERSA provides a general-purpose framework for defining and recognizing events in live or recorded surveillance video streams. The approach for event recognition in VERSA is using a declarative logic language to define the spatial and temporal relationships that characterize a given event or activity. Doing so requires the definition of certain fundamental spatial and temporal relationships and a high-level syntax for specifying frame templates and query parameters. Although the handling of uncertainty in the current VERSA implementation is simplistic, the language and architecture is amenable to extending using Fuzzy Logic or similar approaches.

VERSA's high-level architecture is designed to work in XML-based, services-oriented environments. VERSA can be thought of as subscribing to the XML annotations streamed by a lower-level video analytics service that provides basic entity detection, labeling, and tracking. One or many VERSA Event Monitors could thus analyze video streams and provide alerts when certain events are detected.

# Table of Contents

## Figures

## Tables

# 1. Introduction

## 1.1. Motivation

Wide area video surveillance is becoming a popular tool for deterring crime and prosecuting those involved in criminal activities. In England, there are approximately 4.2 million closed circuit television cameras [1], many thousands of which are employed by law enforcement officials to monitor urban areas. While the coverage is extensive, there is no effective way to monitor thousands of cameras in real-time for suspicious activity. It is too costly and man-power intensive. Because of the lack of real-time monitoring, the extensive surveillance provided by these cameras has been primarily useful in post-incident investigations, and less useful as a means to deter or prevent criminal activity.

What is needed is a computer system that can provide un-blinking, un-tiring, continuous real-time monitoring of a large number of cameras. The ideal system could recognize suspicious or threatening behavior patterns or events, within some reasonable level of confidence, and could alert law enforcement officials upon detection. The technology would act as a sophisticated filter, allowing a dramatic reduction in manpower required to provide the quality of surveillance necessary to deter illicit activity and immediately react to threats or criminal acts as they arise. As an added benefit, the same technology could be applied to search recorded video to assist in post-incident investigations.

Building such a system requires addressing long-standing challenges at many levels of Computer Vision. Some of the typical challenges include: segmentation of objects of interest from complex and dynamic backgrounds, accurately tracking an object from one frame to the next in a video sequence while dealing with occlusions, classification and recognition of an object under arbitrary viewing angles and partial occlusions, and so on. The vast body of research performed over the past several decades has yielded some very good techniques to address these challenges – although all are far from perfect. Some would say that it is an art as much as a science to determine what combination of techniques will yield the best results for a given situation.

Over the past few years, commercial products have emerged that claim to provide intelligent monitoring of surveillance cameras. Although useful to some extent, the features these products provide are still quite limited. These products often claim to recognize *specific* patterns, such as loitering behavior, but none provide a simple mechanism to define an arbitrary behavior pattern to be detected.

There are, however, mature techniques for tracking moving objects in a video stream, such as Kalman Filters and Mean Shift Trackers. Similarly, supervised learning methods such as Neural Networks, Support Vector Machines, and other methods can be trained to reliably distinguish, for example, pedestrians from automobiles in a parking lot surveillance video stream.

The technology to detect, classify, and track moving objects of interest in a stationary camera’s video stream has developed to the point where higher-level reasoning about the objects is possible. This, in short, is the motivation for this thesis research -- to develop the higher-level capability required to describe and detect a wide array of possible events in fixed-camera surveillance video footage. The technology developed under this thesis is called “*VERSA,”* or *Video Event Recognition for Surveillance Applications*.

## 1.2. Problem Statement

This thesis seeks to address the problem of detecting events in the video stream of a stationary surveillance camera. In particular, this thesis proposes that spatial and temporal logics can be used to match event templates to the video data being analyzed. The event templates can be created by people of moderate technical aptitude and do not require offline training or statistical modeling.

A video stream is a sequence of images, or frames. Each frame contains a set of objects that are detected by a lower-level video analytics routine. Each object in a given frame has a unique identifier and a type (classification). In this video sequence, there are *intra-frame* relationships (those relationships between objects in the same frame) and *inter-frame* relationships (those relationships between objects in different frames).

Many intra-frame relationships are spatial, such as “above,” “near,” “overlaps,” and so on. No attempt is made to determine the actual 3D positional relationships of the objects – extracting 3D information is extremely difficult to do reliably and in real-time from a typical surveillance camera. Instead the spatial relationships that are presented in this thesis are based on the 2-dimensional image plane. The 2D information is sufficient in many cases to generate reasonable Event Templates. The downside is that Event Templates thus generated may be most effective from a particular camera’s pose. Yet many typical use-cases for the proposed technology would require location-specific Event Templates. For example, one may wish to monitor a specific doorway to detect any after-hours usage. The limitation that the spatial reasoning is performed in the 2D image plane does not dramatically impact the usefulness and applicability of the proposed technology.

Many of the inter-frame relationships are temporal, and may be applied between entire frames. For example, an event definition may require that a particular *keyframe* (a frame of video where the objects present match the specified intra-frame relationships) occur "before" another particular keyframe. Many of the temporal relations deal with intervals of time, such as "during."

An event in the video stream is a composition of identified objects and the spatial and temporal relationships between them. An Event Template is a specification of an event using specific predicates and syntax (i.e., the "VERSA language"). Recognizing an event requires matching the Event Template, within some allowed tolerance/uncertainty, to the live video data stream.

To do so, one must specify the spatial and temporal operators of the VERSA language and map the intra- and inter-frame relationships from the video sequence into appropriate symbolic terms. One must be able to recognize an instance of a given relationship and thus be able to combine these instances into the recognition of a given event.

Conceptually the VERSA Language could combine lower level events into higher level interpretations in a manner similar to how complex programs can be created by combining lower level modules. In this manner, with a few basic building blocks, subject matter experts could employ a user-friendly interface to describe, in a concise manner, the type of event they are interested in. The system would be able to automatically "compile" the specification into an Event Template to be uploaded into an intelligent video analysis product to extend its capabilities to recognize this new event definition.

The objectives of this thesis research include:

- Developing or extending a language to allow for concise descriptions of the entities and relationships present in a video sequence. This is the *VERSA Language*.
- Designing a translation interface that converts the output of a lower-level computer vision system (that provides object tracking and classification) into the syntax of the VERSA Language.
- Demonstrating the application of the VERSA Language to create a set of representative Event Templates.
- Building a recognizer that can analyze a video sequence and detect occurrences of events specified as VERSA Event Templates. This is the *VERSA Event Monitor*.

## 1.3. High Level Architecture

Figure 1 provides a diagram of VERSA's high level architecture. The diagram is useful in illustrating the context of the technology developed in this research effort and in delineating the lower level processing mechanisms that are not addressed within the scope of this effort. In Section 2, there is an overview of several technologies that, in whole or in part, address VERSA's low level video processing needs.

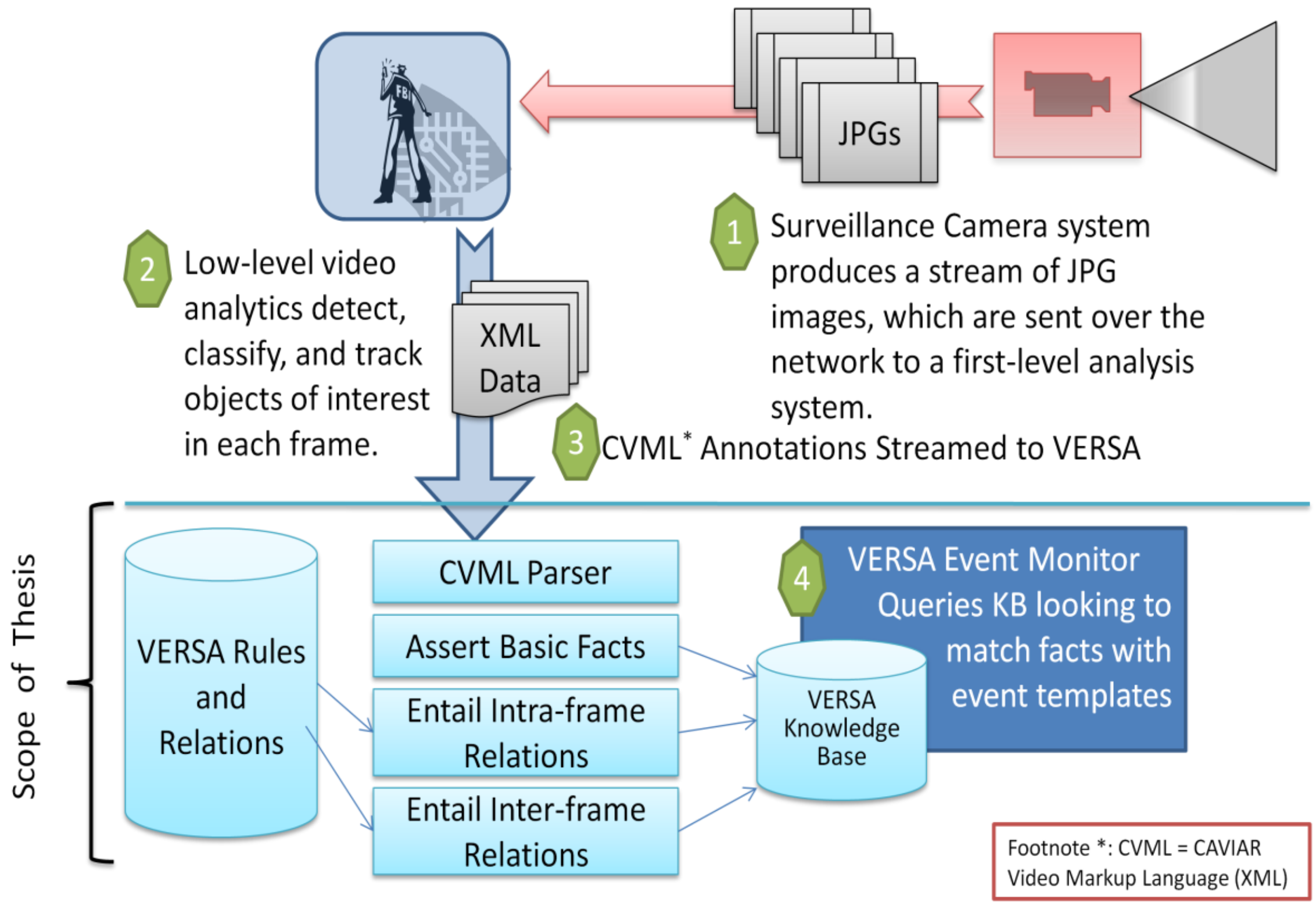

**Figure 1: VERSA High Level Architecture.**

The core components of VERSA include:

- A parser that reads the input data stream and asserts the basic (or *ground*) facts into a knowledge base. The input data stream is assumed to be formatted as XML. In particular, the implementation presented in this thesis uses CVML, the CAVIAR Video Markup Language [2].
- A repository for the definition of VERSA's spatial and temporal relationships and the production rules that are used to assert a relationship given a set of facts.
- A knowledge base of the facts that have been generated from parsing the input data stream and the facts that have been entailed from the production rules.
- A mechanism for monitoring/querying the knowledge base to determine if there is a match to an Event Template.

- An optional graphical user interface (not shown in diagram) that helps users to define and test Event Templates.

# 2. Background

## 2.1. Data Sets

The application domain for my thesis is video surveillance from fixed cameras. A useful data set might come from video recorded at a parking lot, train station, mall, or other public area with pedestrian traffic. Three data sets were identified that are publicly available for the development and evaluation of video analytic systems. They are:

1. NIST TRECVid data set [3]
2. PETS data set [4]
3. CAVIAR project's data set [5]

The NIST TRECVid project seeks to evaluate performance of video information retrieval technologies [6]. Data sets are updated every year, but are generally taken from public broadcast news, documentaries, and the like. The videos used by TREC 2001 are available at the Open Video website [7], but none are surveillance video. Later TRECVid data sets (2003-2006) are archived by the Linguistic Data Consortium (LDC) managed by University of Pennsylvania. Unfortunately, LDC charges $500 for non-members to acquire the video.

The PETS (Performance Evaluation of Tracking and Surveillance) Metrics On-line Evaluation Service project has a dataset from the PETS 2001 workshop that is freely available. A portion of the data set has been labeled by hand. The data is surveillance-oriented. This data set proved to be less valuable due to the format of the annotations and the limited amount of ground-truth annotated video.

The CAVIAR (Context Aware Vision using Image-based Active Recognition) project's goal was to investigate ideas central to a cognitive vision approach. They created a freely available data set which includes video from surveillance cameras in a shopping mall and the lobby area of the INRIA building. The data has been hand-annotated with outlines of the people in the scene, approximate direction of travel, orientation, and more. The annotations are encoded in the CAVIAR Video Markup Language (CVML), which is an XML-based markup language [2]. This was the data set selected for this thesis due to the convenience of the XML annotations and the number of readily available videos on their website.

Because the CAVIAR data is hand-annotated, it provides a level of detail and certainty beyond what is likely to be provided by an automated system. There is information provided in the CVML data that is ignored as being beyond the reliable capabilities of today's video analytics software. Some of that data relates to the interpretation of the video, such as when individual people can be considered a group of people or what activity that person is engaged in. VERSA only makes use of object labels, bounding boxes, and trajectory information provided by the CVML annotations.

## 2.2. Video Annotation Languages and Tools

### 2.2.1. MPEG-7

MPEG-7 is a multimedia content annotation standard [8]. It serves to standardize the format for creating descriptive metadata that is time-linked with multimedia files, such as music or video. MPEG-7 is based on XML. Unfortunately, there hasn't been a wide adoption of MPEG-7 for semantic video annotation, so finding a tool which produces MPEG-7 annotations has been challenging.

VideoAL [9] is described as an end-to-end MPEG-7 video automatic labeling system. Unfortunately, it was confirmed by the author that VideoAL is no longer in development and no source code or binaries are available to other researchers wishing to expand upon the system or reproduce its results.

IBM Research has a tooled called VideoAnnEx, which is freely available [10]. This tool provides assistance in generating MPEG-7 compliant annotations to videos. It is primarily a utility for the hand-annotation of video sequences.

Eptascape, Inc. markets a commercial product that generates MPEG-7 data streams describing the contents of surveillance video and uses an analysis engine on the MPEG-7 data for identifying behaviors and detecting events [11].

### 2.2.2. CAVIAR Video Markup Language (CVML)

Mentioned earlier, CVML is an XML specification for the annotation of video frames. CVML was employed by the CAVIAR project to provide hand-labeled ground truth data for their video data set. This is currently the XML markup used in VERSA, but it would be a relatively minor change in the implementation to parse other XML-based annotation languages.

### 2.2.3. VEML/VERL/EDF

VEML and VERL stand for, respectively, Video Event Markup Language and Video Event Representation Language. Developed by Navatia, Bolles, Hobbs, François, and others ([12] and [13]), these languages were used as a basis of the US Government's Advanced Research and Development Activity (ARDA) 2004 Event Taxonomy Challenge Project to generate an event ontology using the well-known standard Web Ontology Language (OWL). VERL is used to describe event ontologies while VEML provides the markup for specific event instances in a video stream. VEML could be used in conjunction with a lower-level video annotation specification such as MPEG-7. In the final report, the authors indicate that representing VERL requires the full OWL representation, while VEML can be represented using the less expressive but more computationally tractable OWL-DL subset.

A related effort by Natarajan and Nevatia is EDF, the Event Description Framework [14]. EDF differs from VERL in that it focuses on a relational representation of events so that queries can be processed using SQL and can take advantage of the availability of spatial indexes and reasoning capabilities that have been developed by the relational database community.

No effort was made in VERSA to generate output in any "standardized" format. Prolog is used as the knowledge base and query interface, thus VERSA's event description language is represented as Prolog predicates. However, it would be an attractive feature to use an enterprise-strength relational database system for large-scale/long-term storage and for providing easier interoperability with other software systems.

## 2.3. Low Level Systems to Provide Tracking and Classification

There are numerous academic and commercial efforts to provide tracking and classification of objects in surveillance video footage. In 2004, Wang and Maybank [15] published a survey of many of these efforts. This thesis will not duplicate the effort of their survey, but it is worth mentioning a few academic and commercial systems that could potentially be used to generate the input annotations required by VERSA.

One of the oldest and best known systems for the detection and tracking of humans in video is Wren et al's Pfinder developed at the MIT Perceptual Computing Lab in the late 1990's [16]. Although Pfinder was created to be an advanced human-computer interface technology, the same technology is applicable to surveillance tasks as well. Pfinder segments the human in the imagery using a background model built from samples of the scene with no human presence. Pfinder then assumes that large changes between the image and the background model are due to human presence.

Carnegie Mellon University's well-known VSAM project (Video Surveillance and Monitoring) [17] describes the use of different techniques for detecting objects of interest in surveillance video. Their primary technique is a hybrid background subtraction/frame difference method.

Another well-known project that addresses the detection and tracking of objects in surveillance footage is the Reading People Tracker developed by Siebel and Maybank (see [18] and [19]) as part of the European ADVISOR project [20]. An open source implementation of the Reading People Tracker is available [21], but is difficult to compile due to its reliance on the fairly arcane YGL graphics library which emulates

Silicon Graphic Inc.'s graphic library routines under X11. As the project is no longer funded, the code base appears to have been essentially orphaned.

University of Southern California researchers headed by Ram Nevatia and Gerard Medioni have been developing video analysis capabilities for many years. In their current VACE project [22] as well as their earlier MOVER project, they employ a model based method that detects, tracks, and estimates the pose of humans by using an appearance model of the parts of a human body. They also have significant research in event detection, which is discussed in Section 3.

Steiger et al present a system that segments moving objects--namely automobiles and people--in surveillance video using the difference between the current image and a background model [23]. Steiger's system tracks the moving objects and generates MPEG7 annotations of the moving objects and tracks. This information is then used as a basis of simple "trip-wire" style event detection (discussed further in Section 3).

Haoran, Rajan, and Liang-Tien [24] describe a technique to generate MPEG-7 compliant motion trajectory information automatically extracted from sports videos.

Hansen et al describe a system to provide for the automatic annotation of humans in video streams [25]. Their approach uses a codebook-based background subtraction algorithm to detect the humans in a surveillance video. Their annotation provides an estimate on the clothing colors, person's height, and focus of attention (direction of gaze).

As an indicator of the maturity of the techniques used to detect and track moving objects in surveillance video, one might consider the plethora of commercially available products that claim to do so reliably. Some of the higher-end systems marketed towards large integrators and government agencies include products from ObjectVideo [26], Eptascape [11], Cernium [27], and IntelliVision [28] (among others). There is even a growing market of less expensive systems marketed towards a broader market, such as NovoSun's CyeWeb [29], which can be purchased starting at $75. A free trial of CyeWeb is available as well for evaluation.

An informal evaluation of CyeWeb's capabilities in detecting and tracking humans in indoor surveillance video was performed. Qualitatively speaking, it did a good job, meeting expectations for what such a tool should be able to do. Unfortunately, CyeWeb does not currently provide any reasonable interface to allow 3rd party tools to take advantage of its detection and tracking results. A product representative for CyeWeb indicated that an interface SDK was under development.

Clearly, given the strong academic results and the availability of numerous commercial products, it is completely reasonable to develop VERSA such that it operates as a layer above these systems to provide for higher level, logical/symbolic reasoning about the objects present in the video stream. Designed to operate in a services-oriented environment, VERSA should be able to integrate with any system that can provide an XML annotation stream.

## 3. Related Work

For the purposes of this thesis, consider the terms "Event Detection," "Behavior Recognition," and "Activity Description" to be essentially synonymous. They refer to the ability for an automated system to monitor a video stream and recognize aspects of the interactions, temporal and spatial relationships, between the entities detected in the video.

While there is some variance in the terminology and perhaps some slight differences in the connotation, the core technology and approaches to detecting events, recognizing behaviors, and describing activities are the same.

## 3.1. Statistical Approaches

One of the more common approaches to event detection in videos is to use a statistical model, most typically a Hidden Markov Model (HMM) or variant. These techniques require the development of a probabilistic model, which is unlikely to be flexible enough for the detection of a wide array of user-specified events. Instead, they are typically used for the detection within a very small set of possible events/behaviors/activities. Models that require supervised training are especially vulnerable to over-fitting and may be very difficult to apply to a different environment without extensive retraining of the model.

Oliver et al [30] developed both a Hidden Markov Model and a Coupled Hidden Markov Model (CHHM) using a synthetic training mechanism for developing the prior distributions. They show that their models perform well at classifying the behaviors between a small set of possibilities of human interactions. In [31], Makris and Ellis employ Hidden Markov Models and *a priori* route designations to model pedestrian behavior. The VSAM project used a Markov Model trained using low-fidelity simulated events to be able to label the activity observed in the video as one of six possible types: “A Human Entered a Vehicle,” “A Human Got Out of a Vehicle,” “A Human Exited a Building,” “A Human Entered a Building,” “A Vehicle Parked,” and “Human Rendezvous.” Furthermore, they tested their system using synthetic events as well [17].

In 2001 conference paper, Hongeng and Nevatia describe an event as being composed of a number of “event threads” where each thread is performed by a single entity (or “actor”). A single event thread is recognized using Bayesian statistical methods. Higher level event recognition is performed as a temporal constraint satisfaction problem over the constituent threads [32].

Gong and Xiang in [33] present a Dynamically Multi-Linked Hidden Markov Model (MDL-HMM) for interpreting group activities. They compare the MDL-HMM to other HMM variants and claim superior performance in their tests. Niu et al [34] propose a simple statistical model developed from the trajectories of the tracked people for classifying the activity as one of three behaviors. They claim their model avoids the complexities of HMM-style implementations while yielding good results.

Dynamic Bayesian Networks (DBNs) have also been used to describe human activities and behaviors. Park and Aggarwal describe a hierarchical Bayesian Network to describe the poses of various body parts at the lower level and a DBN is used to detect how the poses change over time [35]. Du et al employed a DBN for describing human interaction using two separate scales for representing object motion in [36].

In [37], Intille and Bobick present a method for describing the activities in an American Football game using a multi-agent belief network, which has some similarity to a naïve Bayes classifier. A different type of statistical approach can be seen in the work of Ivanov, Bobick, et al in [38] and [39]. They propose a two-level system. The first level consists of a set of independent probabilistic detectors of basic events. The second level is a stochastic context-free grammar parsing mechanism. The probabilistic grammar is used to address the longer time-line events and to disambiguate between lower level events. Although this approach appears to be more extensible than those previously discussed, it still requires the significant development of both a set of low-level probabilistic event detectors and a probabilistic grammar for the higher level parser.

## 3.2. 2D Geometric Approaches

In 2001, Ayers and Shah developed a system for recognizing a small set of human actions that occur in an office environment [40]. Their approach requires the prior description of the room layout and the definition of certain probabilities regarding whether a part of the scene has changed or not. Bounding boxes and distance thresholds provide much of the context for activity recognition, and thus their approach may be classified as a 2D Geometric Approach. As with most of the Statistical Approaches described above, Ayers and Shah can only recognize a small set of activities.

Many commercial products also provide for the recognition of a fixed set of possible event types, and most of those events are based on simple 2D geometric reasoning. For example, ObjectVideo’s product allows the user to draw boxes or lines and detect when a moving entity has crossed a particular boundary in a particular direction. It also provides modes for the detection of loitering behavior, detecting items left behind by an individual, and for counting individuals in a scene.

CyeWeb provides for the detection of similar event types. Their online manual lists the following: “Detect Object Moving inside Region,” “Detect Object Enter or Leave Region(s),” “Detect loitering object(s),” “Detect objects congregating in an area,”

“Detect illegally stopped objects,” “Detect object cross line(s),” and “Object counter” [41].

VERSA’s flexible syntax and query capabilities allows for the definition of Event Templates to handle more than just a fixed number of event types. Additionally, using VERSA, one can recognize composite events and develop queries that provide for more sophisticated temporal reasoning.

## 3.3. Declarative/Logical/Linguistic Approaches

In addition to statistical and simple geometric approaches, there has been research relating to declarative approaches, both logical and linguistic, for event recognition. In 2000, Rota and Thonnat developed a logical formalism for activity recognition in video streams [42]. Their formalism uses a uniform representation format for both the description and the recognition of the models of the concepts involved in an activity. Given their proposed representation, events are recognized by computing the solution to a Constraint Satisfaction Problem (CSP). However, the published details provide very little information on how they implemented and evaluated their solution. A few years later, Vu, Bremmond, and Thonnat extended this work to improve the recognition speed and allow for longer term events by implementing new interval-based temporal operators (see [43] and [44]). VERSA’s overall approach to event detection is very similar to Thonnat et al’s, but adds additional spatial and temporal predicates, further develops the architecture, and uses Prolog as the uniform representational format.

Katz et al used a linguistic approach to develop a prototype system, called Spot, designed to answer questions about the contents of a video [45]. Although not exactly the same problem domain as real-time event detection, their approach has obvious utility for information retrieval and post-incident investigation on stored surveillance video.

Another interesting technique is presented by Ghanem et al in a 2004 paper [46]. Ghanem uses Petri Nets for representing and recognizing events in surveillance video data. A Petri Net is a mathematic construct originally developed for modeling discrete distributed/concurrent systems. Ghanem presents a query-based system that constructs a Petri Net from the user's query by combining simpler event nets appropriately. Recognition is performed by the tokens which propagate through the network.

The most similar video event recognition system to VERSA is the VidMAP (Video Monitoring of Activity with Prolog) program presented by Shet, Harwood, and Davis in 2005 [47]. As with VERSA, VidMAP resides as an architectural layer above one or more low-level vision algorithms used to generate primitive facts of interest. VidMAP is implemented in Prolog and uses predefined rules evaluated over the primitive facts for detecting events. VERSA provides more spatial and temporal relations, provides more advanced temporal reasoning using interval sets, and provides a more complete implementation. Additionally, VERSA is designed with a services-oriented architecture in mind, includes the specification of how the lower level systems are integrated via XML, and provides a graphical interface to allow the user to "sketch" and test an Event Template. It is also unclear as to how much more VidMAP may rely on lower level systems than VERSA. In their paper, the authors describe that a 15-minute video clip generates 357 facts which are then used by VidMAP to recognize events. In comparison, VERSA would generate many thousands of facts within that same time period, but uses an interval set representation for efficient event recognition.

# 4. Methods

## 4.1. Overview of Approach

VERSA's approach to event recognition combines geometry with declarative logic and temporal reasoning. This section discusses the spatial and temporal reasoning predicates and then provides a discussion of the handling of uncertainty in VERSA.

## 4.2. Spatial Reasoning

In the definition of the spatial relationships discussed in this section, the following variables are used. In all cases, subscripts may be used to distinguish different instances.

- $\tau$ is a frame of the current video being processed.
- $\chi$ is an entity in the video frame. An entity may be of any classification provided by the underlying video annotation language. In this thesis, only two types of entities are considered: people and objects, where an object is any unidentified item.
- $\rho$ is a pixel location, or "point," with a known $x$ and $y$ coordinate on the image.
- $\Gamma$ is a rectangle in the image plane, defined by the four corner points starting with the bottom left and proceeding clockwise.

### 4.2.1. Basic Facts for All Entities

The CVML Parser asserts these fundamental facts for each entity in each frame of video:

- $exists(\chi, \tau)$ : entity $\chi$ exists in frame $\tau$.

- $type(\chi, T, \tau)$: T is the type/class label for entity $\chi$ in frame $\tau$.
- $bounds(\chi, \Gamma, \tau)$ : provides the bounding rectangle for the given entity in the given frame of video.
- $loc(\chi, \rho, \tau)$ : provides the center point of the bounding rectangle of the entity in the frame.
- $orient(\chi, \theta, \tau)$ : θ is the angle representing the "orientation" of the entity in the given frame of video. The angle is specified in CVML such that 0 degrees points straight up, and the angle is read clockwise. θ is typically the direction of motion of the entity. If non-moving, the orientation is not applicable, and $\theta = 0$.
- Also note that each of the three predicates, *bounds*, *loc*, and *orient*, imply *exists*. For example, the assertion $loc(\chi_3, \rho, \tau_{256})$ implies $exists(\chi_3, \tau_{256})$. This implicit relationship is relied upon in the definitions of the spatial and temporal relationships shown later.

### 4.2.2. 2D Geometry for VERSA

VERSA defines the following geometric relationships between points and rectangles to provide a basis for the spatial relationships between entities.

**Table 1: 2D Geometry Point/Rectangle Relationships**

| **Supporting Predicates for 2D Geometry** | **Interpretation** |
|---|---|
| $pt(x, y, \rho)$ | ρ is the point defined by the coordinates *x* and *y*. |
| $rect(\rho_1, \rho_2, \rho_3, \rho_4, \Gamma)$ | Γ is the rectangle defined by the four points $\rho_1$, $\rho_2$, $\rho_3$, $\rho_4$. |
| $ptInside(\rho, \Gamma)$ | The point, ρ, is inside or on the bounds of the rectangle Γ. |
| | Another way to look at this is that the point is a member of the set of all points that comprise the rectangle. $\rho \in \Gamma$. |
| $rectInside(\Gamma_1, \Gamma_2)$ | $\Gamma_1$ is entirely inside bounds of $\Gamma_2$ or has the same bounds. From a set perspective, $\Gamma_1 \subset \Gamma_2$. |
| $overlaps(\Gamma_1, \Gamma_2)$ | Rectangle $\Gamma_1$ overlaps with rectangle $\Gamma_2$, which means they have at least one point in common. From a set perspective, $\Gamma_1 \cap \Gamma_2 \neq \emptyset$. |
| $dist(\rho_1, \rho_2, d)$ | The Euclidean distance between the two points, $\rho_1$ and $\rho_2$, is *d*. |
| $rectHigher(\Gamma_1, \Gamma_2)$<br>$rectLower(\Gamma_1, \Gamma_2)$ | $\Gamma_1$ has a center point that is higher/lower of that of $\Gamma_2$. Nothing is said of their respective horizontal positions. |
| $rectLeft(\Gamma_1, \Gamma_2)$<br>$rectRight(\Gamma_1, \Gamma_2)$ | $\Gamma_1$ has a center point that is left/right of that of $\Gamma_2$. Nothing is said of their respective vertical positions. |
| $maxX(\Gamma, x)$<br>$minX(\Gamma, x)$ | The maximum/minimum x-coordinate of is $\Gamma$ given by *x*. |
| $maxY(\Gamma, y)$<br>$minY(\Gamma, y)$ | The maximum/minimum y-coordinate of $\Gamma$ is given by *y*. |
| $inXrange(\rho, \Gamma)$<br>$inYrange(\rho, \Gamma)$ | True if point $\rho$ has an x/y-coordinate that lies within the range of *x*/*y*-coordinates included in rectangle $\Gamma$. |
| $inXrange(\Gamma_1, \Gamma_2)$<br>$inYrange(\Gamma_1, \Gamma_2)$ | True if $\exists \rho \ni ptInside(\rho, \Gamma_1) \wedge inXrange(\rho, \Gamma_2)$<br>True if $\exists \rho \ni ptInside(\rho, \Gamma_1) \wedge inYrange(\rho, \Gamma_2)$ |

### 4.2.3. VERSA 2D Spatial Relationships

From the fundamental facts asserted by the CVML Parser module and the basic geometry predicates defined above, one can define the following spatial relationships between two entities in a given frame of video.

**Table 2: VERSA Spatial Relationships**

| **Definitions of VERSA Spatial Relationships** |
|---|
| $dist(\chi_1, \chi_2, \tau, d) \leftarrow loc(\chi_1, \rho_1, \tau) \wedge loc(\chi_2, \rho_2, \tau) \wedge dist(\rho_1, \rho_2, d)$ |
| $overlapping(\chi_1, \chi_2, \tau) \leftarrow bounds(\chi_1, \Gamma_1, \tau) \wedge bounds(\chi_2, \Gamma_2, \tau) \wedge overlaps(\Gamma_1, \Gamma_2)$ |
| $inside(\chi_1, \chi_2, \tau) \leftarrow bounds(\chi_1, \Gamma_1, \tau) \land bounds(\chi_2, \Gamma_2, \tau) \land rectInside(\Gamma_1, \Gamma_2)$ |
| $outside(\chi_1, \chi_2, \tau) \leftarrow bounds(\chi_1, \Gamma_1, \tau) \land bounds(\chi_2, \Gamma_2, \tau) \land \neg overlaps(\Gamma_1, \Gamma_2)$ |
| $near(\chi_1, \chi_2, \tau) \leftarrow dist(\chi_1, \chi_2, d) \land (d \geq d_0)$ , where $d_0$ is a threshold distance. |
| $higher(\chi_1, \chi_2, \tau) \leftarrow bounds(\chi_1, \Gamma_1, \tau) \land bounds(\chi_2, \Gamma_2, \tau) \land rectHigher(\Gamma_1, \Gamma_2)$ |
| $lower(\chi_1, \chi_2, \tau) \leftarrow bounds(\chi_1, \Gamma_1, \tau) \land bounds(\chi_2, \Gamma_2, \tau) \land rectLower(\Gamma_1, \Gamma_2)$ |
| $above(\chi_1, \chi_2, \tau) \leftarrow$<br>$bounds(\chi_1, \Gamma_1, \tau) \land bounds(\chi_2, \Gamma_2, \tau) \land higher(\chi_1, \chi_2, \tau) \land inXrange(\Gamma_1, \Gamma_2)$ |
| $below(\chi_1, \chi_2, \tau) \leftarrow$<br>$bounds(\chi_1, \Gamma_1, \tau) \land bounds(\chi_2, \Gamma_2, \tau) \land lower(\chi_1, \chi_2, \tau) \land inXrange(\Gamma_1, \Gamma_2)$ |
| $moreLeft(\chi_1, \chi_2, \tau) \leftarrow bounds(\chi_1, \Gamma_1, \tau) \land bounds(\chi_2, \Gamma_2, \tau) \land rectLeft(\Gamma_1, \Gamma_2)$ |
| $moreRight(\chi_1, \chi_2, \tau) \leftarrow bounds(\chi_1, \Gamma_1, \tau) \land bounds(\chi_2, \Gamma_2, \tau) \land rectRight(\Gamma_1, \Gamma_2)$ |
| $leftOf(\chi_1, \chi_2, \tau) \leftarrow bounds(\chi_1, \Gamma_1, \tau) \land bounds(\chi_2, \Gamma_2, \tau) \land moreLeft(\chi_1, \chi_2, \tau) \land$<br>$inYrange(\Gamma_1, \Gamma_2)$ |
| $rightOf(\chi_1, \chi_2, \tau) \leftarrow bounds(\chi_1, \Gamma_1, \tau) \land bounds(\chi_2, \Gamma_2, \tau) \land moreRight(\chi_1, \chi_2, \tau) \land$<br>$inYrange(\Gamma_1, \Gamma_2)$ |

## 4.3. Temporal Reasoning

Harry Chen's CoBrA project [48] implements a temporal reasoner over ISO 8601 Dates that adheres to the DAML Time Ontology published by J.R. Hobbs [49]. The CoBrA implementation is in SWI Prolog, which made it convenient to adapt for use in VERSA. Chen's source code is provided under the Creative Commons Attribution 1.0 Generic license [50].

Unlike CoBrA, VERSA's representation of time is based on discrete frame numbers. The VERSA adaptation maintains the DAML Time Ontology relationships but removes the ISO 8601 Date formats with integer frame numbers and intervals. The following interval-interval and point-interval relationships over frame sequences have been implemented, mostly thanks to Chen's code.

Following Prolog conventions, "+" indicates an argument must be bound to a ground term, "-" means that it should be unbound, and "?" means that it could be bound or not. The arguments are list structures, which may represent a single point in time (an "instant") such as *[923]* for frame 923, or an "interval" such as *[923, 958]*. A "proper interval" is one where the first element in the list is strictly less than the second.

### 4.3.1. Temporal Data Types

- *instant(+TL)*
- *interval(+TL)*
- *proper_interval(+VL)*

### 4.3.2. Instant-Interval Relationships

- *begins(?BL,+TL)*
- *ends(?BL,+TL)*
- *before(+XL,+YL)*
- *after(+XL,+YL)*
- *inside(+IL,+VL)*
- *begins_or_in(+IL,+VL)*
- *time_between(+VL,+IL1,+IL2)*

### 4.3.3. Interval-Interval Relationships

- *int_equals(+VL1,+VL2)*
- *int_before(+VL1,+VL2)*

- *int_meets(+VL1,+VL2)*
- *int_met_by(+VL1,+VL2)*
- *int_overlaps(+VL1,+VL2)*
- *int_overlapped_by(+VL1,+VL2)*
- *int_starts(+VL1,+VL2)*
- *int_started_by(+VL1,+VL2)*
- *int_during(+VL1,+VL2)*
- *int_contains(+VL1,+VL2)*
- *int_finishes(+VL1,+VL2)*
- *int_finished_by(+VL1,+VL2)*
- *starts_or_during(+VL1,+VL2)*
- *nonoverlap(+VL1,+VL2)*

#### 4.3.4. Interval Sets

One distinguishing characteristic of VERSA's temporal reasoning is the use of Interval Sets. An Interval Set is simply a set of intervals. Having the capability to represent time as Interval Sets can be useful when trying to represent sets of non-contiguous intervals. One may wish to reason about how two Interval Sets relate to each other. For example, suppose that a certain relationship holds over a set of various time periods. Specifically, suppose that $near(\chi_1, \chi_2, \tau)$ is true for a number of frames F. One can determine all frames where this is true by using the Prolog *findall* predicate as follows:

```
findall( F, near(id1, id2, F), F).
```

The result might look something like:

```
F = [0, 1, 2, 3, 4, 5, 6, 7, 8|...]
```

where the list containing the F values is too long to show all values, hence the ellipsis at the end. With Interval Set logic, one can coalesce the results using the query:

```
findall( F, near(id1, id2, F), F), make_iset(F, Fset).
```

where the predicate *make_iset* generates the canonical Interval Set (as defined by J. Paine [51]) out of the list of frames. The result now looks like the following:

```
Fset = [0--437, 446--450, 511--516]
```

Now one can easily see the temporal intervals (frame intervals) where this relation holds and that the intervals are not contiguous. This format is clearly easier to read, but also provides opportunities to find specific intervals within the Interval Set that satisfy other temporal interval-interval relationships such as *int_before*.

VERSA extends Paine's Prolog Interval Set implementation [51]. Paine's Interval Set logic includes many of the same interval reasoning predicates as the CoBrA code, yet operating over interval *sets* instead of single intervals. VERSA adds predicates to make it easy to construct canonical Interval Sets from lists of frame numbers, to provide for a set of relations that work on the interval members of the Interval Sets, and to provide a conversion between Paine's interval notation and that used by Chen's implementation. Paine's Interval Set code defines a custom Prolog operator, denoted by "--", that represents a single interval. In Chen's CoBrA implementation, the interval *T1--T2* would be written as a 2-element list *[T1,T2]*.

VERSA provides a general purpose predicate that looks for the existence of a specific interval-interval relationship between the members of two Interval Sets. The

*find_intervals(+Functor, +ISet1, +Iset2, -Interval1, -Interval2)* predicate provides this capability and is defined in Prolog as follows:

```
find_intervals( Functor, ISet1, ISet2, I1, I2) :-
    member(A1--B1, ISet1),
    member(A2--B2, ISet2),
    I1 = [A1,B1],
    I2 = [A2,B2],
    call(Functor, I1, I2).
```

One should note the use of the *call* predicate that invokes the specified functor (which is the name of one of the interval-interval predicates previously defined) given the two intervals found as members of the two Interval Sets. Those familiar with Prolog will notice that this predicate will return a single solution to the query. Should there be multiple solutions among the intervals that comprise the two interval sets, they can be found by invoking the Prolog *findall* (or similar) meta-query predicate.

### 4.3.5. Timestamp Lists

It is often convenient to represent certain query results in a special list structure created for VERSA called a *timestamp list*. A timestamp list is one in which each list member consists of a key-value pair where the key is an identifier and the value is the timestamp represented as a frame number.

An example of a timestamp list is the following, which might represent all the frames where a particular entity (labeled a, b, c, or d) satisfies some condition, such as being near a doorway.

```
[a-14, a-13, a-12, b-27, a-99, a-100, b-50, c-15, c-16, c-29, d-100]
```

A timestamp list does not require that the elements be sorted. Members with the same identifier do not have to be consecutive in the list, nor do the timestamps for the same identifier have to be listed in order. However, it is often convenient to sort and/or group the results by the key values, so VERSA provides predicates to do so.

VERSA also provides a way to represent a timestamp list using Interval Sets, which often results in a more compact notation and can be used to more easily reason about durations where some condition holds true. Given the timestamp list presented above, the following is the format when converted into an Interval Set notation.

```
[a-[12--14,99--100],b-[27--27,50--50],c-[15--16,29--29],d-[100--
100]]
```

## 4.4. Reasoning with Uncertainty

### 4.4.1. Interval Set Smoothing

VERSA relies on an underlying video analytics package to provide object detection and tracking. As these systems are not 100% reliable, there may be brief periods of time where an object is not detected, misclassified, or lost by the tracker. At the VERSA level, these momentary lapses would present themselves as gaps, or *fragmentation*, of the Interval Set where a particular relationship holds true.

VERSA employs a temporal smoothing operator applied to Interval Sets that can be used to address this fragmentation. The operator could be implemented in many ways, but one simple way is to apply a 1D version of the *closing* morphological operator to an Interval Set represented as a bit stream. The morphological closing operator is a composite operation that first performs morphological *dilation* followed by *erosion*. The intent is to "fill-in" small gaps in the pixels of a binary image. In the dilation phase, every 1 bit expands such that any neighboring 0's become 1's as well. In the erosion phase, the 1's that neighbor 0's are erased. The net effect is that small gaps are filled-in while large gaps remain unchanged.

Consider the notional example shown in Table 3. Let *A* represent an Interval Set where some fact is true at the frames indicated by reading horizontally left to right in the first row of the table. In this example, *A* is the Interval Set {3-4, 6-9, 14-16}. After applying the dilation operator, *A* is transformed into Interval Set *A'* = {2-10, 13-17}. The third row completes the closing operation by applying erosion to *A'* to yield *A''* = {3-9, 14-17}. As one can see in this example, the 1-frame gap at frame 5 in *A* has been filled in, while leaving the contiguous interval 14-16 unchanged.

**Table 3: Example of Interval Set smoothing using the closing operator.**

| | **1** | **2** | **3** | **4** | **5** | **6** | **7** | **8** | **9** | **10** | **11** | **12** | **13** | **14** | **15** | **16** | **17** | **18** |
|---|---|---|---|---|---|---|---|---|---|---|---|---|---|---|---|---|---|---|
| **A** | | | 1 | 1 | | 1 | 1 | 1 | 1 | | | | | 1 | 1 | 1 | | |
| **A'=d(A)** | | 1 | 1 | 1 | 1 | 1 | 1 | 1 | 1 | 1 | | | 1 | 1 | 1 | 1 | 1 | |
| **A''=e(A')** | | | 1 | 1 | 1 | 1 | 1 | 1 | 1 | | | | | 1 | 1 | 1 | | |

### 4.4.2. Fuzzy Logic

Another area of uncertainty is the terminological vagueness of certain relationships, such as the concept of nearness. Fuzzy Logic is an approach for reasoning about such terminological uncertainty. Lacking any Fuzzy Logic, one might define a threshold distance for the *near* relationship[1]. If the distance between two entities is less than this threshold, they are considered near, otherwise they are not. The problem with such a "crisp" threshold is that there is a sharp state change over an almost imperceptible 1-pixel shift in relative object positions.

---

[1] The logical definition of the near relationship as shown earlier in Table 2 is a "crisp" definition and does not make use of Fuzzy Logic.

In contrast, Fuzzy Logic allows a system to provide a level of truth to a given proposition. The *near* relationship, for example, might be unequivocally true if the entities in question are overlapping, but as the distance between them increases, the truth value becomes increasingly divided over the propositions of *near* and *not near*, until after some distance *not near* is unequivocally true. If one continues increasing the distance, perhaps there is another overlap between *not near* and *far*.

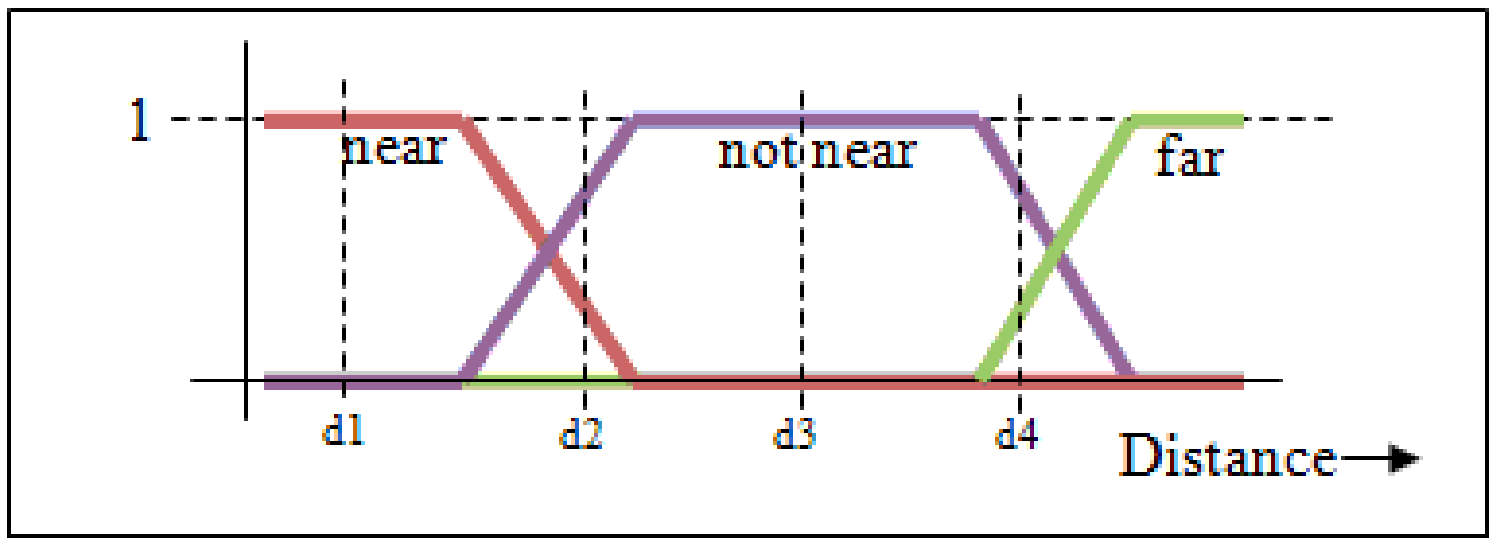

**Figure 2: Fuzzy Logic Illustration of Nearness Relationship**

In Figure 2, the *x*-axis represents the relative distance between two entities and the *y*-axis represents the fuzzy truth value (varying between 0 and 1) assigned to each term. The overall nearness concept consists of the terms *near*, *not near*, and *far*. At distance d1 the entities in question are unequivocally near to each other. At distance d2, however, there is some uncertainty with most of the truth value being given to *not near*, but with some overlap to *near*. As the distance increases to point d3, one becomes very confident that the distance is best described as *not near*. As the separation continues to increase between the entities, one enters the transition between *not near* and *far*, and the finally the entities may become unequivocally *far* from each other.

There have been a number of different implementations of Fuzzy Logic reasoning using Prolog over the past twenty years. In the mid-1980's, Ishizuka and Kanai discussed incorporating Fuzzy Logic in Prolog-ELF [52]. A Master's thesis by Bradley Richards around the same time describes his Fuzzy Logic Prolog implementation [53]. Martin, Baldwin, and others developed a fuzzy-logic version of Prolog -- also in the 1980's -- which eventually became known as FRIL, or Fuzzy Relational Inference Language [54]. In the mid-90's, a Fuzzy Logic Prolog using Łukasiewicz implication was proposed by Yasui et al [55]. And more recently, Guadarrama et al implemented a variant of Fuzzy Logic Prolog using real-domain constraint logic programming [56]. Fuzzy Logic and Prolog are a natural fit, and Fuzzy Logic reasoners have been implemented in Prolog for nearly as long as Prolog itself has been around. Although the current VERSA implementation does not use Fuzzy Logic predicates, it would be a worthwhile extension to the system for future research.

### 4.4.3. Match Score

Another way to deal with uncertainty is to allow for partial matches of a given frame or event template in the query results. In the definition of a frame template, one specifies the entities that must (and must not) exist in the frame and the relationships between those entities that must be true. To allow for partial matches, one can also specify how many of the relationships must hold (either as a fixed number or as a percentage) for a given frame to match a defined template.

Allowing partial matches in this manner makes a trade-off between sensitivity and selectivity. While partial matches can be used to "cast a wider net" (reduction of false negatives, or not recognizing an event when it occurs), they have the downside of generating more false alarms (increasing false positives, or recognizing an event when it actually has not occurred.)

VERSA currently provides a query mechanism for using partial matches in frame templates. This same mechanism can be set to disallow partial matches by simply setting the match percentage to 100. (See Section 5.3. for more on the query syntax as implemented.)

# 5. Implementation

## 5.1. SWI-Prolog

VERSA is based on logical reasoning about the spatial and temporal relationships between the entities in the source video stream. As such, Prolog seemed to be an ideal choice as the programming language used to implement much of the VERSA technology. Prolog is a popular declarative logic programming language. Although there is an ISO standard for the Prolog language, there are a number of popular variants available to choose from.

SWI-Prolog is a free open source Prolog language distribution originally created by Jan Wielemaker [57]. SWI-Prolog was selected over other Prolog implementations because of the richness of SWI-Prolog's packages, including solid support of XML and Semantic Web markup languages and easy integration with Java.

## 5.2. CVML Parser

CVML (CAVIAR Video Markup Language) [2] was selected as the input video annotation language primarily because of the availability of several ground-truth annotated surveillance videos freely available from the CAVIAR project website [5]. CVML is based on XML, so it would be relatively easy to refactor this aspect of VERSA to use other XML-based annotation languages as input.

For basic XML parsing, VERSA uses SWI-Prolog's SGML/XML Parser package known as "sgml2pl" [58]. The sgml2pl parser generates a hierarchical list data structure in Prolog that represents the XML data, but in a format that makes it easy for Prolog predicates to query the data. A representative sample of CVML for two frames of a video is presented in Appendix A. The predicate *load_xml_file(+Filename, -XMLdata)* of the sgml2pl package provides a simple way to parse the entire contents of an XML file and unifies the *XMLdata* variable with the resulting list data structure. This approach works well when working offline with an annotated video recording, such as the CVML files provided for the CAVIAR data sets. A slightly different method would have to be used for parsing a real-time XML annotation stream.

Once the XML has been parsed into the list data structure, then VERSA CVML parsing predicates are used to assert facts into the knowledge base. An important predicate in doing so is *processFrame(+XMLdata, +FrameNum),* which causes the basic facts for the specified frame to be asserted into the knowledge base and also entails the spatial relationships that hold for the entities in the frame, asserting those relations as facts into the knowledge base as well. The essential code[2] of the *processFrame* predicate is shown below. Note that in Prolog, comments are indicated using the % symbol, and are shown here in italic font.

[2] I say "essential" because the code has been simplified in this presentation for clarity, most notably by removing certain aspects of the code that have more to do with Prolog's operational semantics than the predicate-logical semantics of the program.

```
%process the annotations for a single frame of video
processFrame(XMLData, N) :-
   framedata(XMLData,N,FrameData), %Extract the frame data
   assertObjects(FrameData,N), %Assert basic facts about each entity
   srFunctors(Functors),  %Get the list of all spatial relationships
   assertRels(Functors,N). %Assert the spatial relations that hold
```

The predicate *framedata(+XMLData, +FrameNum, -FrameData)* is used to parse the information for a specific frame from the CVML data structure. Because this predicate must recursively search both depth and breadth of the XML data structure, it is a bit more complex in implementation.

```
%We are done if X is not a list
framedata(X,_,_):- atomic(X),!.

%Simple case when the XMLData starts with the frame element we want.
framedata([element(frame,[number=Na],Data)|_],N,Data):-
   atom_number(Na,N),!. %N is the numeric vale of Na

%breadth search once we've found the 'frame' level in hierarchy to
%find the frame number of interest
framedata([element(frame,[number=Ma],_)|Rest],N,Data) :-
   atom_number(Ma,M),
   M < N,  %not this frame...keep looking for siblings...
   framedata(Rest,N,Data).
framedata([element(frame,[number=Ma],_)|_],N,_) :-
   atom_number(Ma,M),
   M > N, !. %somehow we're beyond the desired frame, stop search.
```

```
%depth search, find if the frame is a descendent of this node
framedata([element(X,_,Rest)],N,Data) :-
   X \= frame,
   framedata(Rest,N,Data).

%breadth search to skip non element() list elements
framedata([X|Rest],N,Data) :-
   X \= element(_,_,_),
   framedata(Rest,N,Data).
```

Given the frame data from the CVML annotation, the *assertObjects(+FrameData, +N)* predicate will parse and assert the basic entity facts for each entity present in the frame data. This is the second predicate invoked by the *processFrame* predicate.

```
assertObjects(FrameData, N):-
   getObjects(FrameData,[Obj|Rest]), %parse the entity list
   assertObj(Obj,N),  %assert info about the first entity in list
   assertObj(Rest,N). %assert info about the rest of the entities

%if frame has empty object list, assert nothing.
assertObjects(_,_).

%helper predicate for assertObjects
%do nothing if X is not a list structure
assertObj(X,_) :-
   atomic(X), !.

%skip the first element in the list if it's not an object element
```

```
assertObj( [Junk|Rest],N) :-
   Junk \= element(object,_,_),
   assertObj(Rest,N).

%the first element is an entity, so process it and then do the rest
assertObj( [element(object,[id=ID],Data)|Rest],N):-
   getBox(Data,box(center(X,Y),Size)),
   getOrient(Data,O),
   getFirstHypothesis(Data,H),
   getType(H,T),
   atom_number(ID,IDn), %IDn is the numeric representation of ID
   assertObj(obj(IDn,T,box(center(X,Y),Size),O,N)),
   assertObj(Rest,N).

%Given an object structure, assert some individual fundamental facts
% that are directly generated from the object structure and useful
% in further processing.
assertObj(obj(IDn,Type,box(center(X,Y),Size),Orient,FrameID)) :-
   assert(loc(IDn,pt(X,Y),FrameID)),    %basic fact: center location
   getRect(box(center(X,Y),Size),Rect),
   assert(bounds(IDn,Rect,FrameID)), %basic fact: bounding box
   assert(orient(IDn,Orient,FrameID)), %basic fact: orientation
   assert(exists(IDn,FrameID)),    %basic fact: exists
   assert(type(IDn,Type,FrameID)). %basic fact: type of entity
```

Once the basic facts for each entity in the frame have been asserted into the knowledge base, the next step is to determine which spatial relationships hold between any pair of entities in the frame and assert those relationships as facts. The list of spatial relationships that are considered are a subset of those listed in Section 4.2., as some are easily derived from the others. This step of the process is accomplished by the *assertRels(+SRFunctors, +FrameNum)* predicate. The *SRFunctors* variable must be unified with a list of those spatial relationships that are to be checked.

```
%assert all the spatial relationships in this single frame.
%do nothing if the functors list is empty.
assertRels([], _).

%process the first functor in the list, and then process the rest.
assertRels([Funct|Functors], Frame) :-
   functor(Term, Funct, 0),
   findall( (ID1,ID2), apply(Term,[ID1, ID2, Frame]), Pairs),
   %pairs are all pairs of ids where this relation holds
   assertRelsAux(Funct, Pairs, Frame),
   assertRels(Functors,Frame).

%do nothing if there are no pairs where this relation holds true
assertRelsAux(_, [], _).

%prevent reflexive relationships. We don't care that an object
%overlaps with itself, for example.
assertRelsAux(Funct, [(ID,ID)|Rest], Frame ) :-
   %duplicate IDs are a no-go, so skip...
   assertRelsAux(Funct, Rest, Frame).

%assert this relationship between these two entities into KB
%Note: we tag "_kb" onto the end of the functor name so that we can
```

```
%differentiate between checking a fact as stored in the KB vs.
%computing the relationship of the same name.
assertRelsAux(Funct, [(ID1,ID2)|Rest], Frame) :-
   concat_atom([Funct, '_kb(',ID1, ',', ID2, ',', Frame,')'],A),
   atom_to_term(A,Term,_),    %might be like near_kb(p1,p2,142)
   assert( Term ),            %add fact to kb
   assertRelsAux(Funct, Rest, Frame).
```

Having completed the processing of the *processFrame* predicate, the CVML parser has completed the task of transforming the annotation into a set of basic facts and intra-frame spatial relationships that reside in the Prolog knowledge base.

## 5.3. VERSA Query Syntax

### 5.3.1. Simple Queries

The querying capabilities of VERSA are implemented in Prolog, so one can use standard Prolog syntax combined with the VERSA predicates to query the knowledge base. The querying capabilities of the system may also be extended by developing additional predicates as deemed necessary or convenient.

The following are some very simple examples of querying the knowledge base, which may be illustrative for those less familiar with Prolog. It is useful to note that in Prolog, variables start with a capital letter, and predicates and constants do not. A second point is that Prolog answers queries by finding the first solution for appropriately unifying the variables in the query. As such, Prolog returns a single answer and must be explicitly told (via user interaction or by using a special predicate) to look for more answers.

**Table 4: Examples of Simple VERSA Queries**

| **Query** | **Interpretation** |
|---|---|
| `near(ID1, ID2, 135).` | Find an entity ID1 that is near an entity ID2 in frame 135. |
| `near(ID1, p1, 135).` | Find an entity ID1 that is near the entity identified as p1 in frame 135. |
| `loc(p1, PT, 56).` | Find the center point PT of the entity p1 in frame 56. Note that the structure of a point is: pt(X,Y). |
| `dist(ID1, ID2, D, 15), D<50.` | Find two entities (ID1 and ID2) whose distance D is less than 50 pixels in frame 15. Note that the comma in Prolog is a conjunction operation. |
| `bounds(o1, RECT, F).` | RECT will be the bounding rectangle for entity o1 in frame F. Assuming the frame data is asserted sequentially, Prolog will respond with the first frame in which o1 exists and will unify RECT to be the bounding rectangle of o1 in that frame. |
| `findall(F, above(p1,p2,F), Fs)` | The *findall* predicate is a Prolog built-in predicate that automatically finds all solutions to a query. In this case, the base query is above(p1,p2,F), and the result of the findall query is to return a list of all frames where p1 is above p2. (Per the definition of *above* in Section 4.2.) |

More advanced VERSA queries are shown later in this section. Before looking at those, however, an important point about querying the knowledge base versus reasoning about the truth of a relationship must be made.

### 5.3.2. Querying the Knowledge Base vs. Reasoning about a Relationship

There is an important distinction, especially in terms of query performance, between a *fact* stored in the knowledge base and the evaluation of a predicate. For example, the *near* predicate defines what it means for two entities to be near each other in a particular frame. The logical definition was presented earlier in Section 4.2, and for illustrative purposes, here is the Prolog implementation used in VERSA.

```
near(ID1,ID2,F) :-
   dist(ID1,ID2,D,F),
   ID1\=ID2,
   threshold(T), %threshold can be set by user
   D < T.
```

The *near* predicate uses a threshold distance, as previously discussed. The threshold distance is found by looking for the first *threshold* fact in the knowledge base and unifying T with that value. For example, if the fact `threshold(50)` existed in the knowledge base, then 50 pixels would be the threshold used for the *near* relation.

When a user issues a query such as `near(ID1, ID2, 135)`, as was shown in the first row of Table 4, Prolog will invoke a search for a solution to the near predicate by first looking for a solution to the first clause, `dist(ID1, ID2, D, 135)`, then the second clause `ID1\=ID2`, and so on, using backtracking when it fails to find a solution for the current variable bindings. This is what is meant when the phrases “reasoning about a relationship” or “searching for a solution” are used in this thesis.

In this situation, however, there is no need to search for a solution if the knowledge base already contains all the facts about the "nearness" of all entities in frame 135. When VERSA parses the frame annotation, as discussed in Section 5.2, it asserts not only basic facts about each entity in the frame (such as the entity's location, bounding box, etc.), but also facts about the spatial relationships between every pair of entities in the frame. In the case of this example, there may already be facts in the knowledge base similar to the following:

```
near_kb(p1,p2,135).
near_kb(p1,o1,135).
near_kb(p4,p5,135).
```

Therefore, a much faster way of finding two entities that are near each other in frame 135 would be to issue the following query: near_kb(ID1, ID2, 135). To make it easy to differentiate between searching the knowledge base for facts about the near relationship and reasoning directly about the near relationship using more primitive facts, VERSA tags the "_kb" extension to the name of the spatial relationship when asserting a fact about a specific instance of that relationship into the knowledge base.

This approach yields a significant query speedup (see 6.3.3. for details). The evaluation of the spatial relationships is done once, when a given frame is being processed by the CVML Parser module. From that point on, the knowledge base can be queried directly, thereby saving the repeated evaluation of the Prolog predicate to search for a solution. In other words, VERSA caches the solutions to the spatial predicates as applied to the entities in each frame of video, so that queries become lookups instead of searches. Section 6 provides an example of the speedup in query performance resulting from this optimization.

As with every performance enhancement, there is a trade-off to caching some of the spatial relationships in this manner. Consider that the cached version of the *near* predicate will have been evaluated with whatever the *then-current* value of the *threshold* fact was. Should you wish to find entities which are near each other using a different threshold value, then the relations stored in the knowledge base will not be useful. The user must set a new threshold value into the knowledge base and then use the *near* predicate to search for a solution. This is no worse than what the user would have done if there were no caching, except for the additional storage used.

Example queries shown in the rest of this section will use the cached relationships whenever possible. Any predicate whose name ends in "_kb" is the cached version of the predicate with the same root name.

### 5.3.3. Using Static Entities in Queries

It is often useful to have a notion of a *static entity* for use in defining Event Templates and other queries. A static entity represents something in the scene of the video that does not change with time. It is assumed present in all frames. An example of a static entity might be a doorway, a kiosk, or simply an area of interest in the video scene.

The *assertStaticElement(+ID, +Box, +Orient)* predicate is used to assert information about a static entity into the knowledge base. The *Box* variable is a structure that provides the static entity's center point and size, as *box( center(x,y), size(width,height))*. The Prolog implementation is shown below.

```
assertStaticElement(ID, Box, Orient) :-
   ground(ID),
   ground(Box),
   ground(Orient),
   %we specify the object with type=static, frame=_, which indicates
   %that the facts we assert about this are true for all frames
   Obj = obj(ID, static, Box, Orient, _ ),
   assertObj(Obj). %assert the basic facts about this static entity
```

Once a static entity has been added to the knowledge base, it is available for queries and can be treated like any other person or object. For example, consider a static entity named “storefront” that has been asserted into the knowledge base as follows.

```
assertStaticElement(storefront, box(center(255,175),
   size(220,40)),0).
```

By issuing the following query, one can find all frames where any person is near the store front. In the first line of the code below, the *findall* predicate finds all the frames where this is true and puts them into the resulting list *Fs*. The second line of code uses the make_iset predicate to convert the list of individual frames, Fs, into the Interval Set variable called *Frames*.

```
findall(F, (near(storefront, ID, F),type(ID,person,F)),Fs),
make_iset(Fs,Frames).
```

The result of this query might look something like the following, which indicates that in frames 12 through 148 and again in frames 162 through 460, there is at least one person near “storefront”.

```
Fs = [12, 13, 14, 15, 16, 17, 18, 19, 20|...],
Frames = [12--148, 162--460]
```

A similar query might be used to detect loitering behavior in front of the store front. This query differs from the previous example in that both the frame number and the entity ID are noted in the result set, so that later one can see if any single person stayed near the store front for a significant duration.

```
findall(ID-F,(near(storefront,ID,F),type(ID,person,F)),Tstamps),
iset_tsl(Tstamps, Res).
```

The *Res* variable is unified with the final query results, as shown in the sample output below. The results are in the Interval Set timestamp list format as discussed in Section 4.3.5. From the results, person “0” was near the store front from frames 12 through 148, person “1” was near it from 177 through 422, and finally person “2” was near it from 162 through 460.

```
Tstamps = [0-12, 0-13, 0-14, 0-15, 0-16, 0-17, 0-18 |...],
Res = [0-[12--148], 1-[177--422], 2-[162--460]]
```

If one had a rule that loitering is defined as a person being near a door for longer than some threshold duration, one could use the results shown above to compare the durations in the timestamp list to the threshold value.

### 5.3.4. Frame Matching

#### *5.3.4.1. Frame Signatures and Frame Templates*

In VERSA, every frame can be described with a *frame signature*. The frame signature provides a listing of the entities that exist in the frame and the type of those entities {person, object}. A frame signature is somewhat like a function signature in a traditional programming language and can be used to find all frames where a specific set of entities exist. The syntax of a frame signature is as follows:

```
frame_sig(FrameNum, TypeList)
```

Where *FrameNum* is a frame number and *TypeList* is a list of elements where each element is a valid type atom followed by a colon and then an entity identifier. The following is an example of a frame signature.

```
frame_sig(98, [object:o1, person:p1, person:p2])
```

Another construct that is important in VERSA is the *frame template*. A frame template is used in "match" queries (discussed later) to find frames that match the key characteristics of the given template. A frame template specifies a *TypeList* (as defined above) except that entities will be variables instead of constants. The frame template also specifies a set of intra-frame relationships that must hold true among the entities in the frame. A frame template also may specify a list of entities that must NOT exist in the desired frame. Finally, a frame template is given a unique identifier so that it can be used in the composition of Event Templates. The frame template predicate is defined as:

```
frametemplate( FrameID, TypeList, RelationsList, NotExistsList).
```

The following is an example of a frame template:

```
frametemplate( ft1, [object:O1, person:P1], [near_kb(O1,P1),
more_left_kb(O1,P1)],[]).
```

The VERSA Graphical User Interface prototype, described in Section 5.5., automatically generates frame templates based on a "sketch" provided by the user.

#### *5.3.4.2. Finding Matches for a Frame Template*

The `frametemplate` structure presented in the previous section provides a convenient way to specify the key entities and relationships to use in finding candidate matches in the video stream for a given key frame. VERSA provides a query predicate called `match` that will return frame matches for a given `frametemplate` specification.

It may be worth noting that the `match` and `frametemplate` mechanisms are provided mostly as a convenience to the user as well as for supporting the VERSA graphical user interface. An informed user could query the Prolog knowledge base and format the results in any way he/she sees fit – in fact, the extensibility provided by Prolog is a key advantage of VERSA's architecture over similar systems using a proprietary language.

There are a few variants of the `match` predicate implemented in VERSA. A description of each is provided below. Each summary includes at least one example showing the correct usage in the SWI-Prolog interpreter. Note that the interpreter's prompt is "?-", and that the interpreter's response to the query is indicated with italics and starting with a ">" prompt. These examples are run against the CAVIAR data set's "Left Bag 1" video, each frame of which is assumed already parsed using VERSA's CVML Parser. The following `frametemplate` has already been asserted into the knowledge base:

```
frametemplate(f1, [object:O1, person:P1], [near_kb(O1,P1)], [])
```

1. Match frames using a `frametemplate` structure. The user provides the identity of a `frametemplate` already asserted into the Prolog knowledge base and a Match Score threshold (see Section 4.4.4). Alternatively, the user provides a `frametemplate` structure in-line with the same predicate. The query will return the first frame that matches. Other matching frames can be found by asking the Prolog interpreter for another solution or by using a predicate such as **findall** or **bagof** to return multiple results at once.

Examples:

```
?- match( Frame, f1, 0.85).
> Frame = 945

?- findall( Frame, match(Frame, f1, 0.85), Fs), make_iset(Fs, Iset).
> Fs = [945, 946, 947, 948, 949, 950, 951, 952, 953|...],
> Iset = [945--972, 1314--1354]

?- match( Frame, frametemplate(_,[object:O1, person:P1],
[near_kb(O1,P1)], []), 0.85).
> Frame = 945,
> O1 = 4,
> P1 = 3
```

2. Match frames without using a **frametemplate** structure. This predicate requires the user to provide all the information that might otherwise be provided in a **frametemplate**, so it is essentially equivalent in operation to the previously described version. This variant is sometimes more convenient to use in queries where a **frametemplate** fact has not previously been asserted.

Example:

```
?- match(Frame, [object:O1, person:P1], [near_kb(O1,P1)], [], 0.85).
> Frame = 945,
> O1 = 4,
```

```
> P1 = 3
```

3. Return an interval set of matching frames by calling the **`iset_match`** predicate. The user must still supply a **`frametemplate`** and Match Score, but the query encapsulates the Prolog code needed to find all solutions and present the result in the interval set format.

Example:

```
?- iset_match( f1, 0.85, ISet).
> ISet = [945--972, 1314--1354]
```

4. Return an interval set of matching frames using the **`iset_match_bindings`** predicate. This differs from the previous technique in that it returns *the associated entity bindings for each interval set*. In the second query shown below, note that a **`findall`** style query is required to return all the Interval Sets and associated entity bindings, and the result is formatted in an Interval Set Timestamp List structure, as discussed in Section 4.3.5.

Examples:

```
?- iset_match_bindings(f1, 0.85, Iset, Entities).
> Iset = [945--972],
> Entities = [object:4, person:3]

?- findall( (E-Iset), iset_match_bindings(f1,0.85,Iset,E), Results).
> Results = [[object:4, person:3]-[945--972], [object:4, person:5]-
[1314--1354]]
```

### 5.3.5. Event Detection

Event Detection in VERSA is a matter of querying the knowledge base using the provided spatial and temporal operators. Those proficient in Prolog may even extend the capabilities by defining new relationships and query mechanisms. VERSA provides a limited mechanism for defining Event Templates (queries) using a graphical user interface as a kind of query-by-example (QBE). VERSA's graphical user interface is discussed in Section 5.5. The Event Templates generated by the VERSA GUI rely on the construction of `frametemplate` structures, as presented earlier, with inter-frame temporal relationships generated based on how the user sequences the frame templates. Although there are many useful Event Templates that can be defined using the GUI, the current implementation does not provide for the full expressiveness of the underlying VERSA language. Examples of the Event Templates generated by the GUI are provided in Section 5.5.4.

One example for detecting a simple event using the VERSA language is shown below. This is an extension to the code discussed in Section 5.3.3. on using static entities in a query. This code detects when any person has remained near a storefront for too long a period of time (500 frames).

```
findall(ID-F, (overlapping(storefront,ID,F), type(ID,person,F)),
   Tstamps),

iset_tsl(Tstamps, Res),

member(ID-Iset, Res),

member(Start--End, Iset),
```

```
End-Start > 500.
```

To make this code into a reusable query, one can write it as follows and save it to a Prolog file.

```
% sample event detection of loitering activity. Area is the id of a
% static entity asserted into the KB. Duration is the duration, in
% frames, that is considered "too long" for a person to be
% overlapping with the area of interest.
% loitering_in(+,+,-,-,-)
loitering_in(Area, Duration, ID, Start, End) :-
   findall(ID-F, (overlapping(Area,ID,F), type(ID,person,F)),
      Tstamps),
   iset_tsl(Tstamps, Res),
   member(ID-Iset, Res),
   member(Start--End, Iset),
   End-Start > Duration.
```

After defining this query, applying it is simply a matter of invoking it periodically as new information is added to the knowledge base by the CVML Parser. The following is an example of the results of applying this query to the CAVIAR Shopping Store Front video.

```
?- loitering_in(storefront, 500, ID, Start, End).
> ID = 1,
> S = 263,
> E = 1066
```

As another example, consider detecting when a person abandons a carried object, as when a person leaves a backpack on the ground and then walks away. One might define this event using three key frames. First is the detection of an object very near to a person. Using that frame as the "existential anchor," one can look for a previous frame where the person exists, but the object does not. Note that one cannot search for the lack of existence of a particular object before one has identified the object in question. In Prolog terms, one must ground the variable that represents the object before checking the *not_exists* condition. The third key frame is one where both the object and person are still in view, but the person has moved far away from the object.

```
%Simple event template for detecting when a person leaves behind an
% object such as a bag or backpack. P is the id of the person, O is
% the id of the object, F_Anchor is the frame where we first see the
% the person near the object.
%left_item(-,-,-).
left_item(P, O, F_Anchor) :-
   match(F_Anchor,[person:P, object:O],[near_kb(P,O)],[],1.0),
   match(F_Prior,[person:P], [], [O], 1.0),
   match(F_After,[person:P,object:O],[not_near_kb(P,O)],[],1.0),
   before(F_Prior,F_Anchor), after(F_After,F_Anchor).
```

The **left_item** query, as defined, can result in many trivial variations of the same answer. For example, it will be true for potentially several frames after F_After, where the person is still far away from the object, but has not yet left the scene. One could refine the code to help eliminate this issue or use the Prolog built-in **once** predicate to find exactly one solution. Testing this query on the CAVIAR Left Bag sample video produces the following results.

```
?- once(left_item(P,O,F)).
```

```
> P = 3,
> O = 4,
> F = 945
```

## 5.4. VERSA Event Monitor

The VERSA Event Monitor is a task that periodically queries the knowledge base to determine if there is a match to any of the active Event Templates. From an architectural perspective, the Event Monitor could be implemented as a separate multithreaded service on the network. It maintains a query connection to the knowledge base and a list of Event Templates. As a network service, multiple Event Monitors could be spawned for load balancing purposes and active Event Monitors could be instructed to change what they are looking for by adding or removing Event Templates from their lists.

Event Monitors could also maintain a set of rules specifying the actions to be taken when an event is detected. Reasonable actions might be to alert the user/watchstander, submit a request to cue the interesting frames of video from the DVR repository for playback, log the event information to a database, trigger a security alarm, etc.

## 5.5. VERSA Graphical User Interface

A graphical user interface (GUI) for VERSA was created as a proof-of-concept to demonstrate how Event Templates might be defined graphically and subsequently used to detect events in a surveillance video stream. It is important to note that the GUI interface is more restricted in the Event Templates it can generate than what could be done with a command line query interface or text editor. However, an Event Template can be exported from the GUI and then tweaked by a knowledgeable user to add additional detection logic.

### 5.5.1. Software Platform

The VERSA GUI is built using Java JDK 6.0 and SWI-Prolog's JPL package [59] for interfacing Java with Prolog. The software was compiled and tested using Microsoft Windows Vista operating system on a laptop with an AMD dual-core processor running at 1.8 GHz and 2 GB of RAM.

Both SWI-Prolog and Java are available for many different operating systems, but it is unclear in the documentation for JPL whether non-Windows operating systems are supported. No attempt was made to run the VERSA GUI on any platform other than what is listed in the previous paragraph.

The icons in the toolbar were adapted from Ken Saunders icon sets [60] available under the *Creative Commons Attribution-ShareAlike 2.5 License*.

### 5.5.2. VERSA GUI Overview

The following figure illustrates the key sections of the VERSA GUI. Details on each section are provided following the illustration.

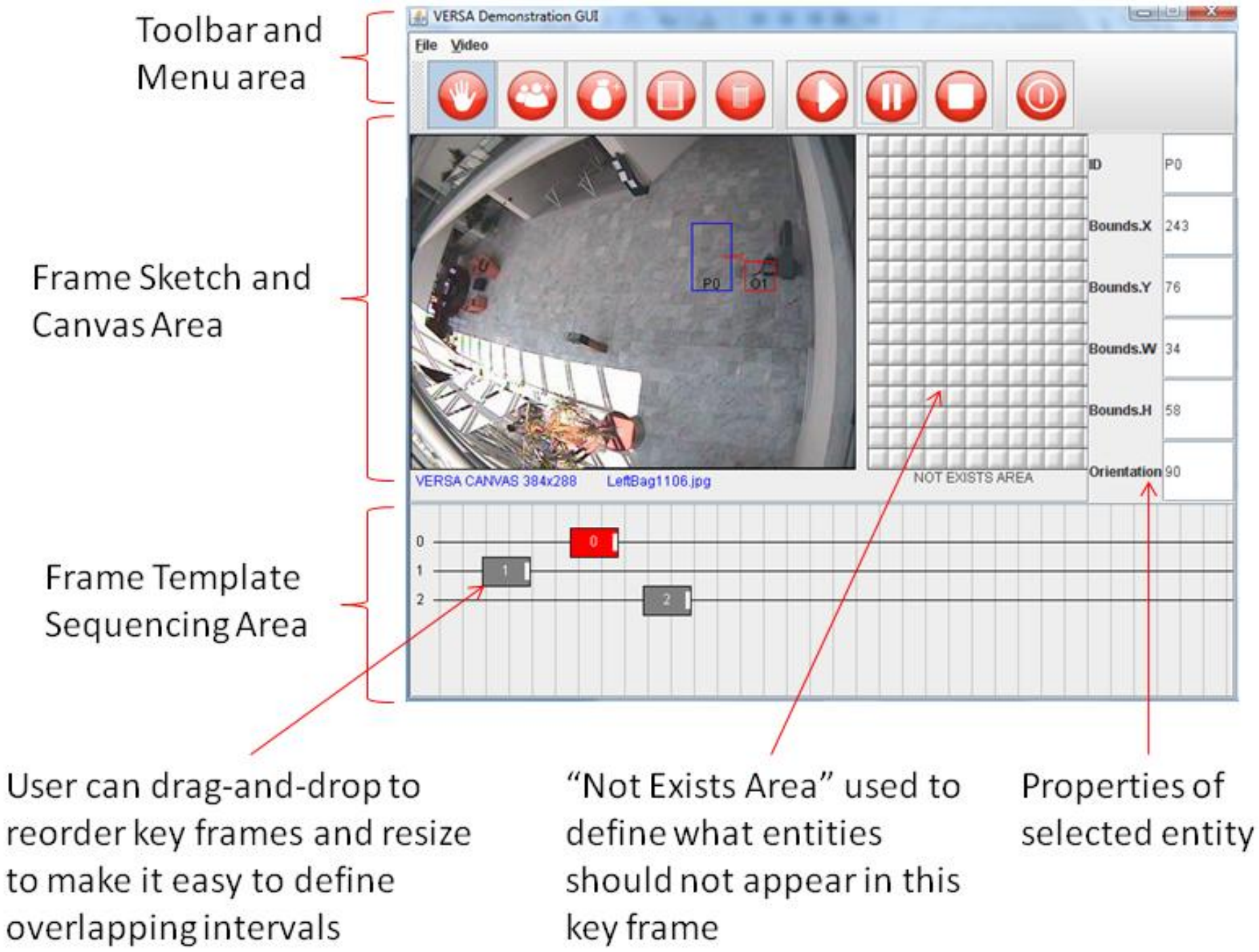

**Figure 3: VERSA Graphical User Interface**

### *5.5.2.1. Toolbar and Menu Area*

The VERSA GUI has two menus, File and Video. The File menu allows the user to load or save an Event Template as well as to export the Event Template as a Prolog query. The Video menu allows the user to set the background, or *canvas* area, to an image extracted from the video stream. The Video menu also allows the user to skip to a particular frame number when testing on pre-recorded video sets.

The icons in the VERSA GUI tool bar are defined as follows:

- 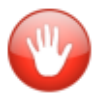 *Selection Mode* – Used to select an entity in the Frame Template canvas, move it, and inspect its properties.
- 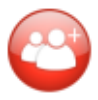 *Person Creation Mode* – Used to add an entity of type "person" to the current Frame Template.

- *Object Creation Mode* – Used to add an entity of type "object" to the current Frame Template.
- *Create New Frame* – Used to add a new Frame Template to the event definition. This will add a new frame to the Frame Sequencing area below the canvas area and clear the current canvas.
- *Delete Selected Object* – Used to delete the currently selected object in the canvas area.
- *Play Video* – Used to play the testing video selected using the Video menu. Each frame of video will replace the canvas backdrop. The current video frame number is shown just below the canvas area.
- *Pause Video* – Used to pause the video playback.
- *Stop Video* – Used to stop the video playback and reset the canvas to the default background. If event detection mode is enabled (see next bullet), stopping the playback will also clear out the Prolog knowledge base.
- *Enable Event Detection* – Used to toggle whether the VERSA GUI is in "event detection" mode. When in event detection mode, the GUI will parse the CVML for each frame of video being played, assert the appropriate relationships into the Prolog knowledge base, and check if the current conditions are a match for the Event Template defined by the user. When not in event detection mode, the video is played without invoking the CVML Parser or checking for matches to the currently loaded Event Template.

### *5.5.2.2. Frame Sketch and Canvas Area*

The Frame Sketch and Canvas Area is where the user defines a single Frame Template for a key frame of video in the Event Template. As previously described, the Frame Template provides the list of important entities in the key frame, including their data types and relative spatial relationships.

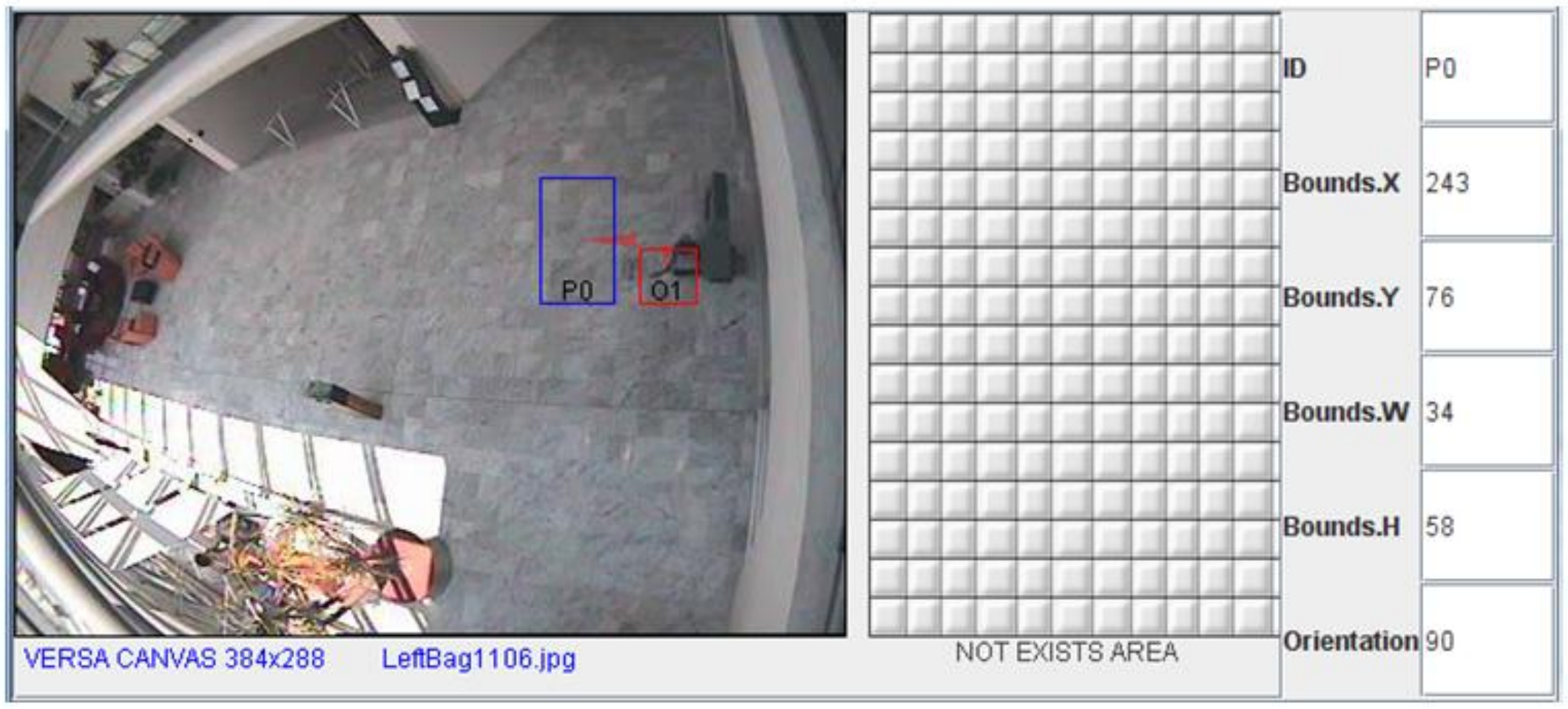

**Figure 4: Close up of Frame Sketch and Canvas Area**

The canvas is the section of the GUI where the video backdrop is displayed and where the user can draw where the entities of interest should be relative to one another and to any static objects defined for the scene. Currently, the GUI supports only two types of entities: People and Objects. People are represented using blue bounding boxes and Objects with red. Information about the currently selected entity in the canvas is displayed on the far right hand side of the GUI, which displays the entity ID, the bounding box location and size, and the orientation.

Between the canvas and the entity properties area is the “Not Exists Area.” This part of the display is used to define those entities that explicitly should *not* appear in this key frame of video. For example, an object which has been picked up and thus should no longer be in view. The user simply creates an object or person with the same ID as one used in a previous Frame Template, and then drags it anywhere in the Not Exists Area. No spatial relationships are inferred by the relative positioning of objects in this special area. Unfortunately, the GUI does not currently provide any way of indicating “forbidden relationships,” or those intra-frame relationships that explicitly should not exist between the entities in the key frame.

Just below the canvas is the video resolution and the current frame of the video being displayed as the canvas backdrop.

In the example shown for a simple Frame Template in Figure 4, there are two entities defined. The entity labeled “P0” is a person facing roughly to the right. The entity labeled “O1” is an object of some sort that is smaller than the person and nearby.

### *5.5.2.3. Frame Sequencing Area*

Below the Frame Sketch area is the Frame Sequencing area, which is used to select a Frame Template from among those used in the Event Template definition and to arrange the Frame Templates relative to one another so that inter-frame temporal relationships can be derived.

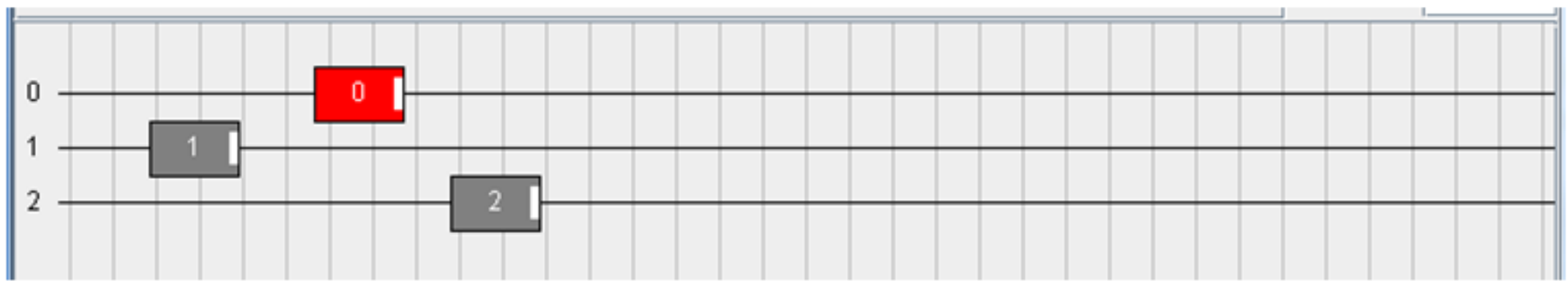

**Figure 5: Close up of Frame Sequencing Area**

In the sample Frame Sequencing Area shown in Figure 5, there are three Frame Templates labeled 0 through 2. The currently selected Frame Template is the one labeled

“0”, highlighted in red. The Event Template defined by this arrangement would look for an interval in which there is a match for Frame Template 1, followed by an interval where there is a match for Frame Template 0, and finally an interval of time following that where there is a match for Frame Template 2. None of the intervals in this example overlap.

The reason one might not have the first Frame Template in the list be temporally first is if one needs to define a key frame for the event where an entity exists *after* a key frame where that same entity explicitly did not exist. Until one defines what the entity is (binds the variable to an instance), there is no way to indicate its lack of existence in an earlier time interval. A typical example would be someone dropping a bag and leaving it. The first key frame likely to be detected is a person near an object. Then the system can look *back* to see if that same person existed in an earlier frame, but not a separate object (because it was being carried and, thus, not detected as a separate entity). At this point, the system can scan future frames to see if the person abandons the object by moving far away from it.

### 5.5.3. Query Generation using the VERSA GUI

A key goal for the VERSA GUI was to demonstrate the feasibility of developing a Query-by-Example mechanism to make it easy for end users to define Event and Frame Templates without having to learn the VERSA language. This section briefly describes how an Event Template query is generated from the “sketch” provided by the user of the GUI.

Essentially, the system looks at what the user has drawn for a given frame as if it were reading CVML annotation, and asserts the basic facts about entities drawn in the Frame Sketch into a temporary Prolog knowledge base, using the previously described JPL interface between Java and Prolog. A unique `frametemplate` ID is used instead of a frame number in the assertions. A *TypeList*, which gives the type and ID of every entity in the frame is created. Then, similar to how the CVML Parser works, a set of spatial relationships between these entities are evaluated, and those that hold true are compiled into the *RelationsList*. Finally, the system checks to see if the user has placed any entities in the "Not Exists Area" of the Frame Sketch. If so, an appropriate *NotExistsList* is created. Finally, a `frametemplate` structure containing this information is returned from the Prolog engine. See 5.3.4.2. for details on the `frametemplate` structure. Because the Prolog code used to generate a `frametemplate` from the GUI sketch is largely the same as that used to assert facts and relationships given a CVML annotation, it is not repeated here.

The Java code must then read the resulting structure and replace all entity identifiers with an associated variable name for use in a query. Doing so involves a string substitution that capitalizes the entity identifier, as Prolog indicates variable names with leading capital letters. This procedure is performed for every frame the user has sketched in the interface, creating an appropriate `frametemplate` structure for each.

To complete the creation of an Event Template, the GUI must generate a set of appropriate inter-frame temporal relationships using the information provided by the user in the Frame Sequencing area of the GUI. Again, using an embedded Prolog engine to entail appropriate relationships, the code converts the left-to-right sequencing of the frames in the GUI as earlier-to-later temporal relationships. The GUI has the capability to entail many of the interval/interval relationships defined in 4.3.3. The following is the Prolog code used to generate the interval temporal relationships from the GUI sketch.

```
%interval relationship functors, used similarly to srFunctors above
irFunctors([int_equals, int_earlier, int_during, int_starts,
   int_finishes, int_meets, int_overlaps]).


% tmp_interval(id,interval) are facts asserted into the kb by the
% versa GUI to be used to construct temporal relationships between
% the frame templates. Each id represents one of the frames in the
% GUI sketch, and the interval has the x-positions of the start and
% stop of the "bar" in the timeline view. Using this information, we
% can determine all of the key temporal-interval relationships.
buildIntervalRels( [], []) :-!.
buildIntervalRels( [Funct|Functs], Rels ) :-
   findall( (ID1,ID2), (tmp_interval(ID1,Int1),
      tmp_interval(ID2,Int2), ID1\=ID2, apply(Funct,[Int1,Int2])),
      Pairs),
   buildIntRelsAux( Pairs, Funct, R ),
   buildIntervalRels( Functs, Relsx ), %add to list structure
   flatten([R|Relsx],Rels).


% converts the interval pairs into a list of temporal relationships
buildIntRelsAux( [], _, [] ) :- !.
buildIntRelsAux( [(ID1,ID2)|Rest], Funct, [Term|Terms] ) :-
   concat_atom([Funct, '(',ID1, ',', ID2,')'],A),
   atom_to_term(A,Term,_),
   buildIntRelsAux( Rest, Funct, Terms).
```

### 5.5.4. Testing an Event Template in the VERSA GUI

The VERSA GUI provides the ability to play back a pre-recorded surveillance video, where the video is stored as a directory of JPG images. Combined with a pre-recorded annotation file in CVML format, this capability allows the user to develop and test Event Templates within a single user interface.

Once the user has defined or loaded an existing Event Template in the VERSA GUI, he can click a toggle button in the tool bar (See 5.5.2.1.) to enable the Event Monitor. When enabled, the toggle button appears pushed-in. As stated earlier, when in event detection mode, the GUI will parse the CVML for each frame of video being played, assert the appropriate relationships into the Prolog knowledge base, and check if the current conditions are a match for the active Event Template. When Event Monitoring is enabled, yellow bounding boxes are displayed over the entities in the video frames as parsed from the CVML annotations. When VERSA detects a match for the active Event Template, information about the query results will be printed to the active Java text console. At this time, there are no visual alerts in the GUI for event detection.

When the toggle button is not engaged, the video is played without invoking the CVML Parser or checking for matches to the active Event Template. Because the CVML annotations are not being parsed when the Event Monitor is disabled, the yellow entity bounding boxes generated from the CVML will not be displayed.

At times, the user may wish to jump to a particular frame number in order to skip uninteresting parts of the video. He may do so by using the Video menu and selecting the Jump To Frame option.

## 5.6. Code Organization

The following table shows the key Prolog modules for the VERSA implementation. In addition to these modules, one may add one or more Prolog files containing sample queries developed using either the VERSA Language or exported using the VERSA GUI.

**Table 5: VERSA's Prolog Modules**

| **Prolog File** | **Module Description** |
|---|---|
| VERSA.pl | Main code module, responsible for parsing CVML and asserting facts about the objects in each frame. Provides the top-level query/reporting interface and the support required by the VERSA GUI. This module includes references to all the others and integrates everything together. |
| VERSA_2D.pl | Simple 2D Geometry relationships. Implements what is described in section 4.2.2. |
| VERSA_Spatial.pl | Spatial relationships used in VERSA, relies on VERSA_2D. Implements relations described in 4.2.3. |
| VERSA_Time.pl | Adapted from CoBrA project, an implementation of the DAML Time Ontology, but using frame numbers and intervals instead of ISO dates. Modified to use the same interval notation as in J. Paine's Interval Set logic and with other adaptations for VERSA. |
| VERSA_ISet.pl | Modified version of J. Paine's Interval Set implementation. Added the *make_iset(+List, -Iset)* predicate and others to support VERSA's needs. |
| svo_util.pl | General purpose utility predicates, such as *unique*(+L), timestamp list sorting/grouping, and other broadly useful predicates. |

## 6. Results

This section presents the application of two Event Templates on a set of six sample surveillance videos, followed by a discussion of performance issues.

### 6.1. Selected Sample Videos from CAVIAR Data Set

As previously mentioned, all sample videos are courtesy of the EC-funded CAVIAR project, freely available at CAVIAR project website [5]. Each video consists of an MPEG-2 encoded video, a folder containing JPGs of each frame, and an XML annotation in CVML format. The following videos were used for developing and testing VERSA.

**Table 6: CAVIAR Videos**

| Video Title[3] | MPEG File | CVML File | Description |
|---|---|---|---|
| Left Bag | LeftBag.mpg | lb1gt.xml | INRIA Lobby, wide angle lens. Person leaves a bag near the kiosk. Other people appear in the scene as well. |
| Left Box | LeftBox.mpg | lbgt.xml | INRIA Lobby, wide angle lens. Person leaves a box near middle of floor and walks away. |
| Meet Crowd | Meet_Crowd.mpg | mc1gt.xml | INRIA Lobby, wide angle lens. Group of four people walking together through lobby area. |

[3] These are my titles, as no official titles are provided on the CAVIAR web site. The file names are CAVIAR's, so one can find the exact video referenced.

| | | | |
|---|---|---|---|
| Shopping Corridor 1 | WalkByShop1cor.mpg | cwbs1gt.xml | Shopping Center in Lisbon. Camera looking down a corridor. Some shoppers enter and exit stores. |
| Shopping Corridor 2 | OneShopOneWait2cor.mpg | cosow2gt.xml | Shopping Center in Lisbon. Camera looking down a corridor. Various shoppers walking towards and away from camera. One shopper enters a store and one waits just outside the entrance. |
| Shopping Store Front | OneShopOneWait2frong.mpg | fosow2gt.xml | Shopping Center in Lisbon. Camera looking across corridor to front entrance of a store. Same scenario as Shopping Corridor but from side view. |

## 6.2. Sample Event Templates

The following two Event Templates are used to illustrate VERSA's ability to recognize events in video streams: *Left Item* and *Loitering*. Each is presented below.

### 6.2.1. Left Item

The following Prolog code, which is a minor modification to that presented in Section 5.3.5., embodies an Event Template for detecting when a person has abandoned a previously carried object.

```
% Simple event template for detecting when a person leaves behind an
% object such as a bag or backpack. P is the id of the person, O is
% the id of the object,F2 is the frame where we first see the person
% near the object (aka the Anchor Frame). F1 is where the person
% first appears on scene. F3 is where the person is no longer near
```

```
% the object
%left_item(-,-,-,-,-).
left_item(P, O, F1, F2, F3) :-
   match(F2,[person:P, object:O],[near_kb(P,O)],[],1.0),
   match(F1,[person:P], [], [O], 1.0),
   match(F3,[person:P,object:O],[not_near_kb(P,O)],[],1.0),
   before(F1,F2), after(F3,F2).
```

The CAVIAR videos used to evaluate this Event Template are the three showing the INRIA building lobby. Sample results are presented below.

### 6.2.1.1. Evaluation on Left Bag Video

The result of executing the `left_item` query on the CAVIAR Left Bag sample video produces the following result.

```
?-left_item(P,O,F1,F2,F3).
> P = 3,
> O = 4,
> F1 = 757,
> F2 = 945,
> F3 = 985
```

The event is detected between entities #3 and #4 in the annotations. Entity #3 is a person and entity #4 is a backpack. The three frames identified by the results are shown below. The yellow and red indicator circles and arrows were added for this illustration.

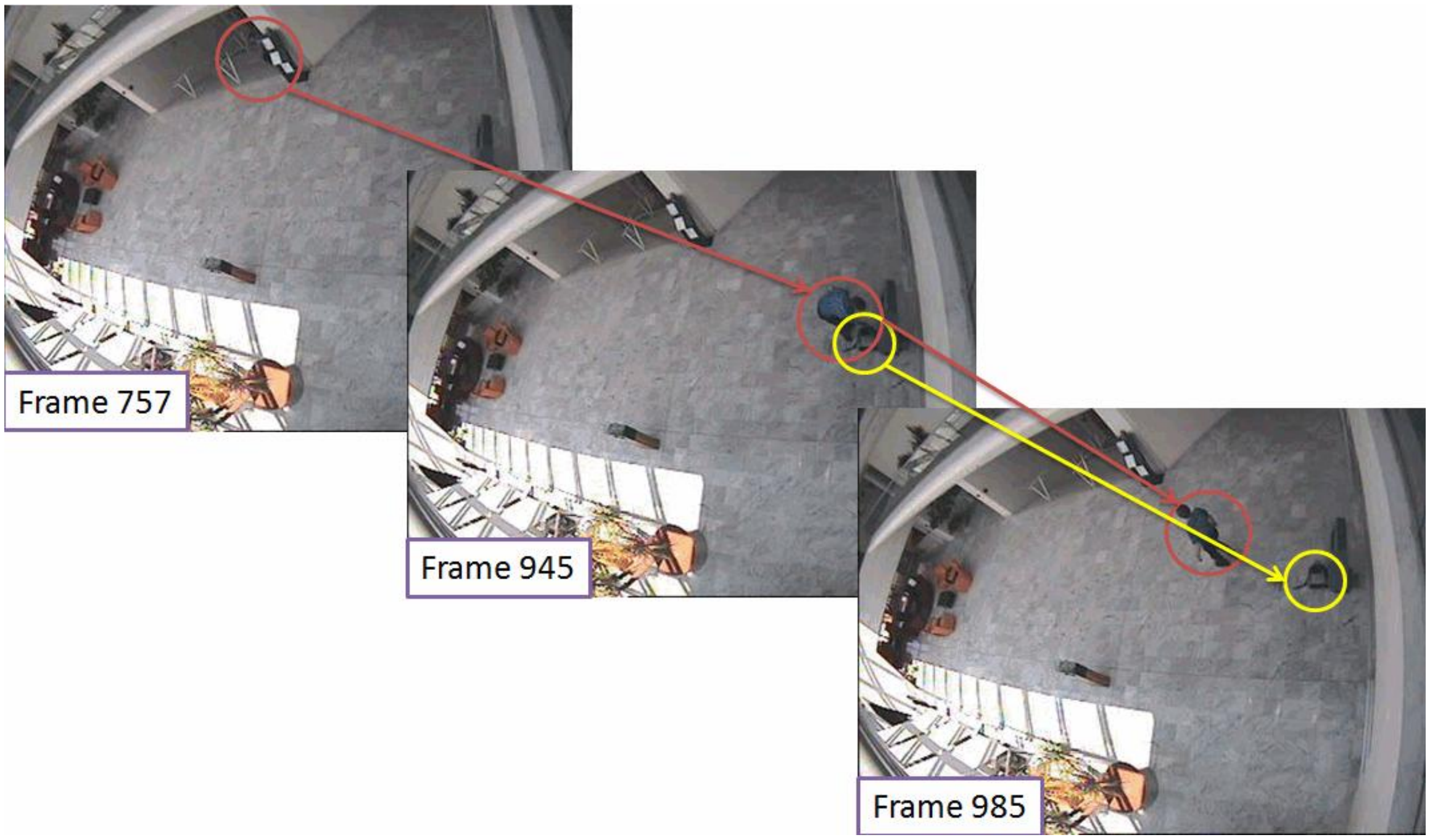

**Figure 6: Left Item Result on CAVIAR Left Bag Video**

#### *6.2.1.2. Evaluation on Left Box Video*

The result of executing the `left_item` query on the CAVIAR Left Box sample video produces the following result.

```
?-left_item(P,O,F1,F2,F3).
> P = 4,
> O = 3,
> F1 = 369,
> F2 = 571,
> F3 = 773
```

The event is detected between entities #3 and #4 in the annotations (coincidentally the same numbers, but different assignment as in the Left Bag video). Entity #4 is a person and entity #3 is a box. The three frames identified by the results are shown below. The yellow and red indicator circles and arrows were added by me for this illustration.

Because the CVML annotations were hand-generated with ground truth data, the earliest frame when an entity appears in the scene is often one where the person is not even fully visible. (See frame 757 of Figure 6, and frame 369 of Figure 7). In an automated system, it is unlikely that the entity would be confidently recognized as a person until several frames later when they are more fully in view. This does not affect the ability of VERSA to detect the event.

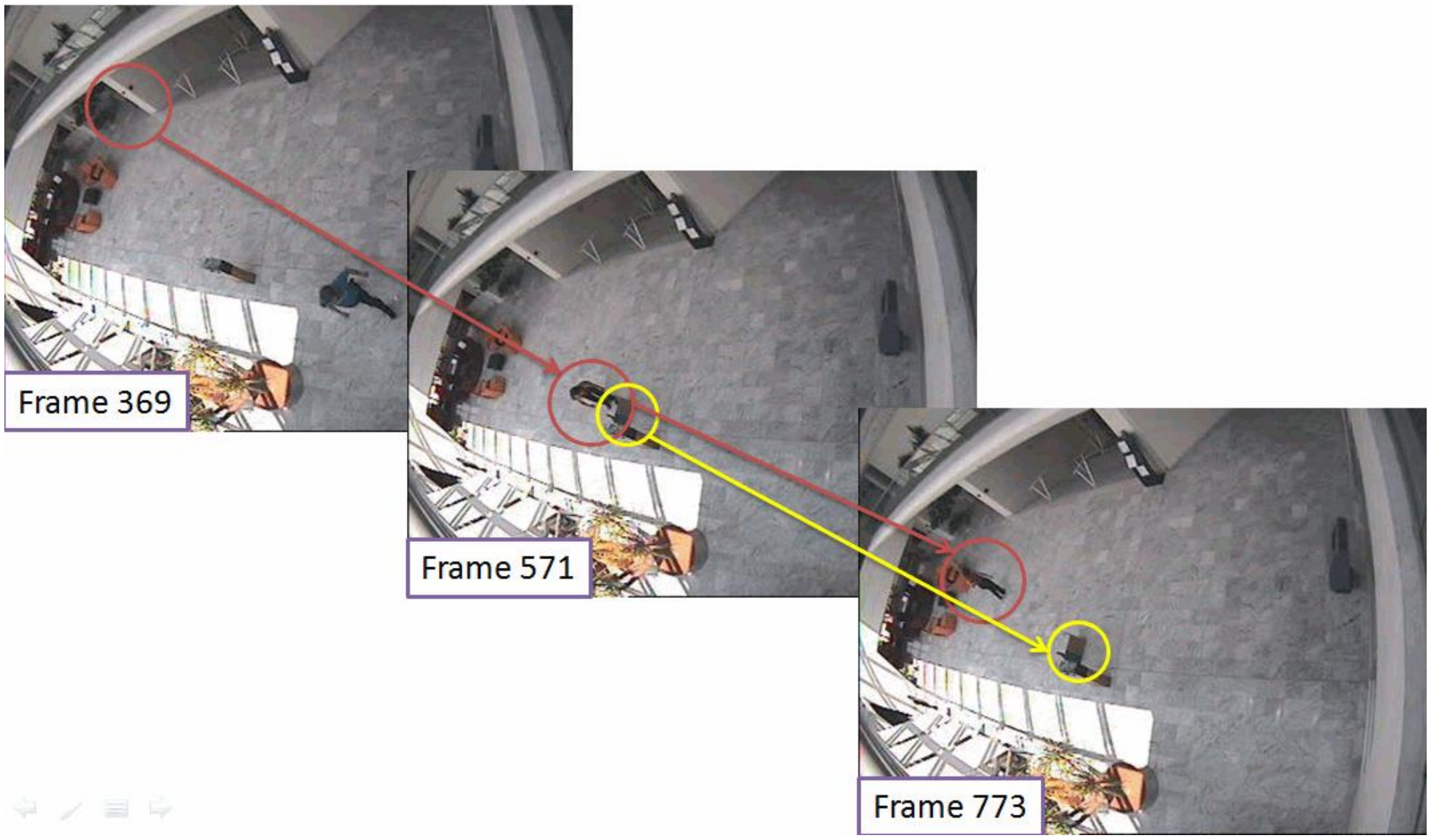

**Figure 7: Left Item Result on CAVIAR Left Box Video**

#### *6.2.1.3. Evaluation on Meet Crowd Video*

This test was done for the sake of completeness. There are no objects left behind in this sample video. Executing the `left_item` query on this video produces, as expected, no results.

### 6.2.2. Loitering

The following Prolog code embodies an Event Template for detecting when a person is loitering in an area of interest defined by the user as a static entity. This code is nearly the same as that presented in Section 5.3.5., but adds Interval Set smoothing as discussed in Section 4.4.1.

```
% sample event detection of loitering activity. Area is the id of a
% static entity asserted into the KB. Duration is the duration, in
% frames, that is considered "too long" for a person to be
% overlapping with the area of interest.
% loitering_in(+,+,-,-,-)
loitering_in(Area, Duration, ID, Start, End) :-
   findall(ID-F, (overlapping(Area,ID,F), type(ID,person,F)),
      Tstamps),
   iset_tsl(Tstamps, Res),
   member(ID-Iset, Res),
   iset_smoothing(Iset,Iset2), %Iset2 has small gaps removed
   member(Start--End, Iset2),
   End-Start > Duration.
```

The following figure illustrates the region used for loitering detection in the Shopping Corridor 1 and Shopping Corridor 2 sample videos, as denoted by the yellow rectangle.

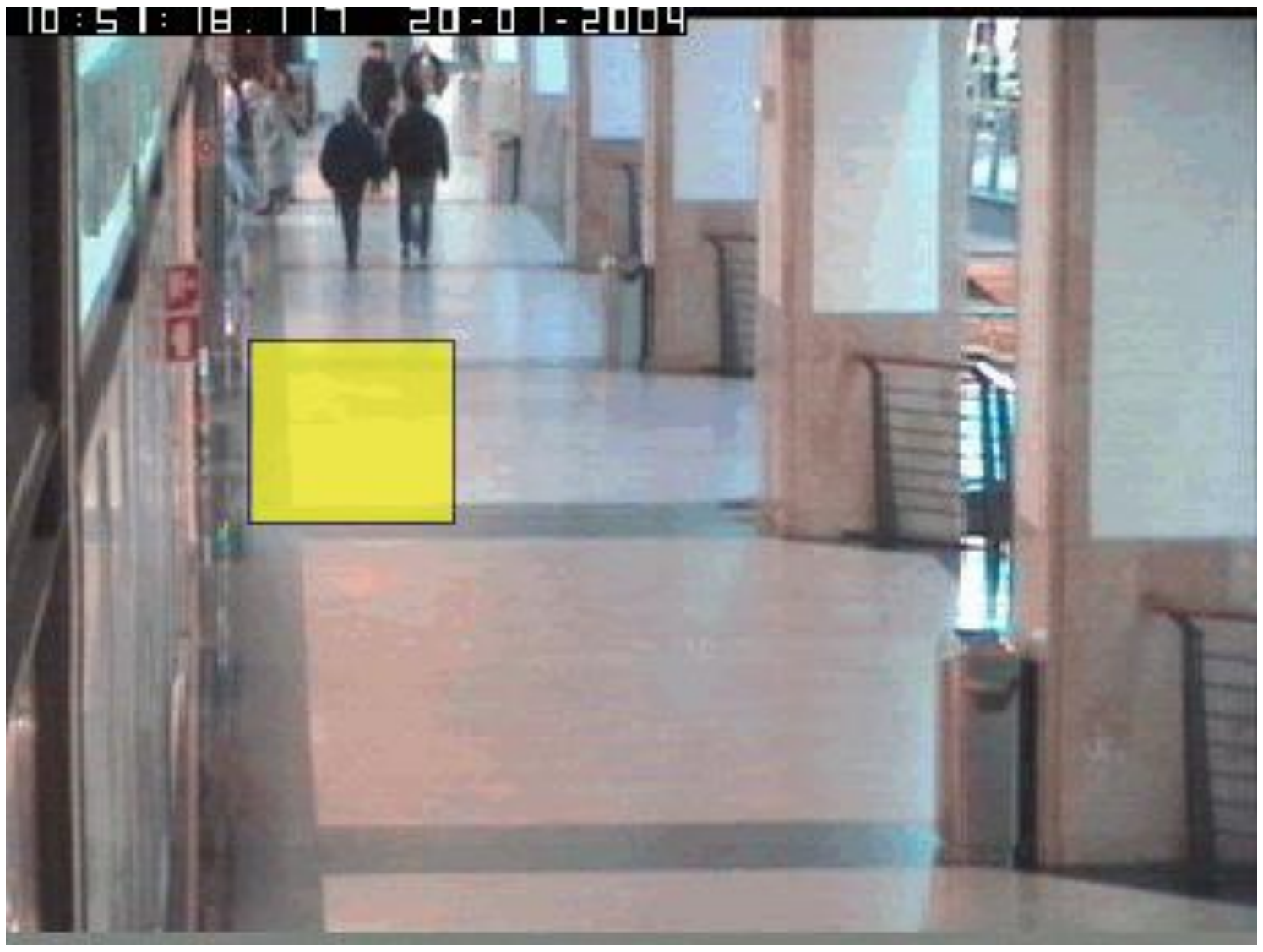

**Figure 8: Area used in Loitering Detection for Shopping Corridor Videos**

#### *6.2.2.1. Evaluation on Shopping Corridor 1*

The following is the result when executing the loitering-in query on the Shopping Corridor 1 video using 500 frames as the minimum duration and the area of interest as defined above.

```
?- loitering_in(area1, 500, ID, S, E).
> ID = 2,
> S = 890,
> E = 1419
```

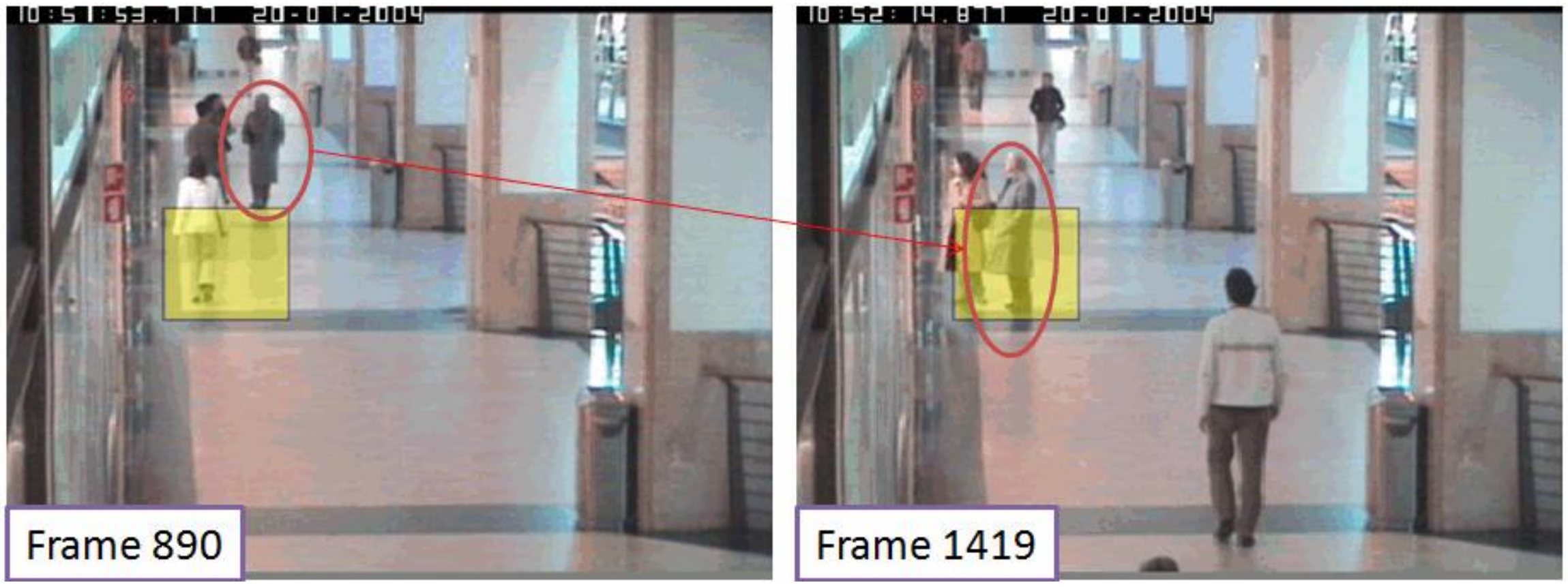

**Figure 9: Loitering Detection on Shopping Corridor 1**

#### *6.2.2.2. Evaluation on Shopping Corridor 2*

When testing the same loitering Event Template with the same area of interest against the Shopping Corridor 2 video, no matches were found. However, when a new area of interest is defined, the following results are produced. In the figure that follows, the area of interest is denoted by the red rectangle. The red triangle within the red rectangle is spurious. It is used by the VERSA GUI to indicate orientation, which is not applicable to a static region.

```
?- loitering_in(area2, 1000, ID, S, E).
> ID = 7,
> S = 149,
> E = 1337
```

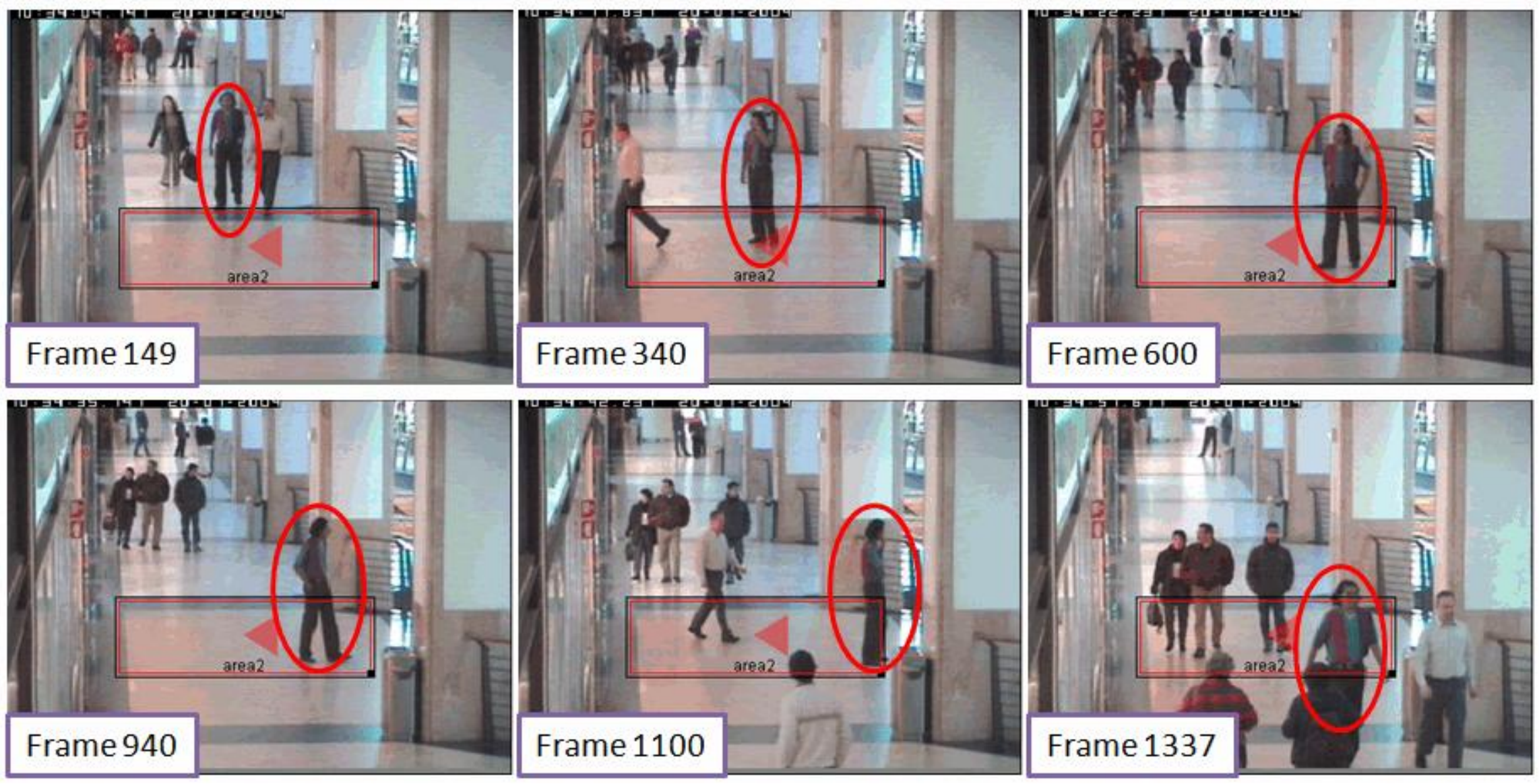

**Figure 10: Loitering Detection on Shopping Corridor 2**

#### *6.2.2.3. Evaluation on Shopping Store Front*

The Shopping Store Front video has a different camera angle than the two videos of the corridor. A new area of interest is defined at the front of the entrance to a store. The following is the result of loitering detection in the Shopping Store Front video.

```
?- loitering_in(storefront, 500, ID, Start, End).
> ID = 1,
> Start = 263,
> End = 1066
```

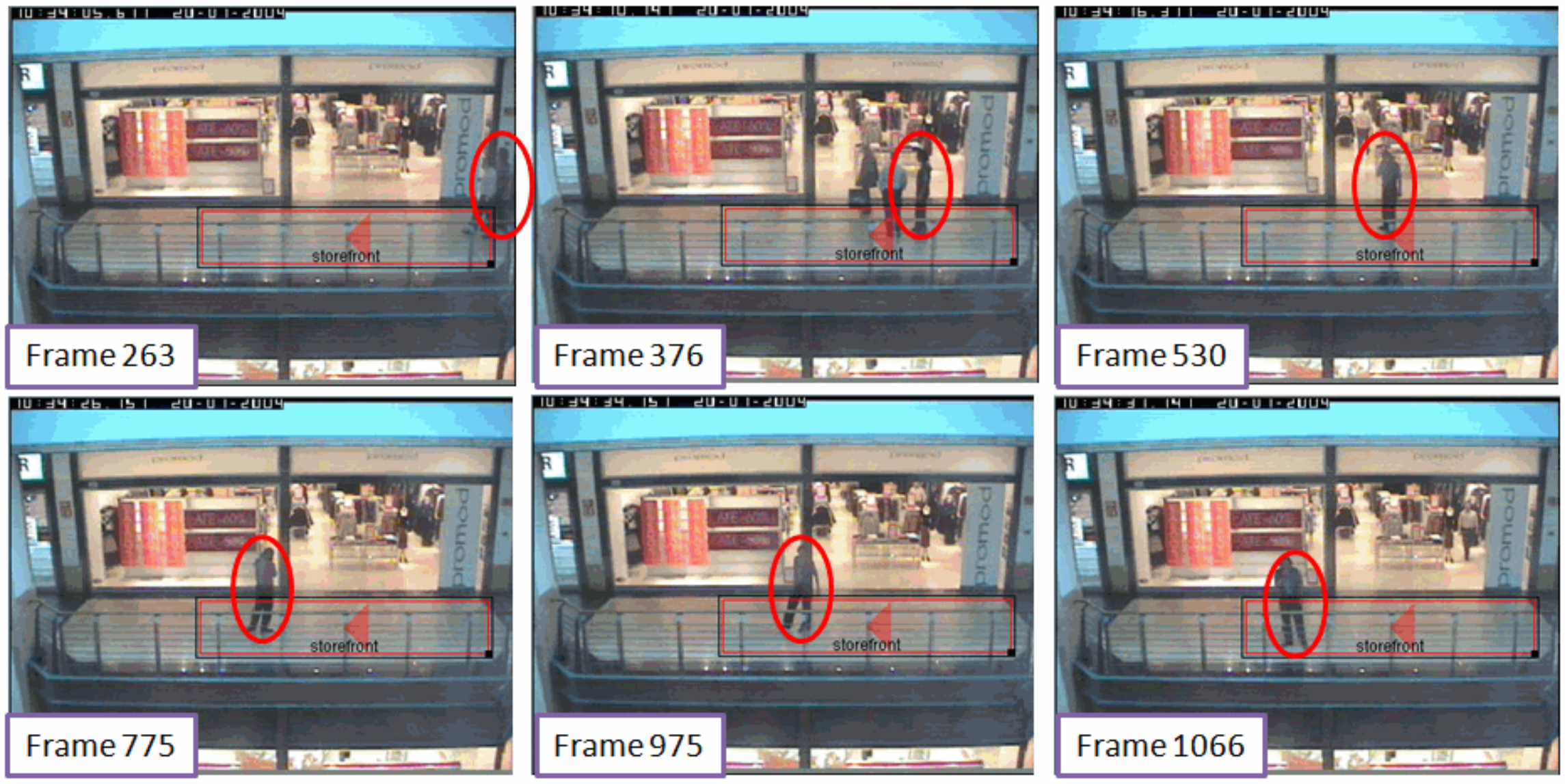

**Figure 11: Loitering Detection Shop Front Video**

## 6.3. Performance Considerations

### 6.3.1. Processing and Query Speed

Table 7 shows the speed at which VERSA can parse a CVML file and assert the basic facts and spatial relations into the knowledge base. The table also shows the typical response time for two representative event queries. All times are given in CPU seconds using the SWI-Prolog built-in `time` predicate. “Processing Time” is the total time to parse the entire video annotation file and assert all facts and relationships into the knowledge base. “Frame Count” is the number of frames in the video. “Average FPS” is the average frames per second computed by dividing the frame count by the processing time. “Average EPF” is the average number of entities per frame, which provides an indication of average scene complexity. “Left_Item Query Time” is the time to get the first match to the Left Item Event Template (see 6.2.1.). Entries with an asterisk (*) indicate that no match was found to the query. “Loitering_In Query Time” is the time to get the first result to the Loitering Event Template (see 6.2.2.). Only the video segments in the Lisbon Shopping center were used with the Loitering template.

**Table 7: Computation Time for VERSA Processing on the Six Videos**

| Video | Frame Count | Processing Time (CPU seconds) | Avg. FPS | Avg. EPF | Left_Item Query Time (CPU Seconds) | Loitering_In Query Time (CPU seconds) |
|---|---|---|---|---|---|---|
| Left Bag | 1439 | 15.57 | 92.4 | 1.59 | 0.41 | - |
| Left Box | 863 | 5.35 | 161 | 1.59 | 0.11 | - |
| Meet Crowd | 491 | 5.93 | 82.8 | 2.37 | 0.14* | - |
| Corridor 1 | 2360 | 522 | 4.52 | 4.81 | Fail** | 1.09 |
| Corridor 2 | 1462 | 285 | 5.13 | 5.18 | 12.3* | 0.53 |
| Shop Front | 1462 | 32.0 | 45.8 | 2.05 | 2.05* | 0.16 |

** The Left_Item query against Corridor 1 did not complete in a reasonable amount of time and was aborted.

In looking at the results presented in Table 7, it is clear that as the scene complexity grows, as measured by the average entities per frame, that the processing throughput (frames per second) is significantly impacted. Because the spatial relationships are typically binary relations, for N entities in a frame, N*(N-1) assertions may be made, ignoring the fact that some relations are symmetric. So the number of assertions will grow on the order of $O(N^2)$. The number of assertions will also increase linearly with the number of frames. Query time when no match is found for an Event Template gets worse as the size of the knowledge base grows, as it must exhaustively check all possible frames.

Clever query design using Interval Sets and Interval Set Timestamp Lists can mitigate some of this effect, but ultimately more needs to be done to improve query speed for production real-time surveillance systems. One possible approach for improving query speed is to apply constraint propagation methods to limit the domain of the search space, as done with CLP (Constraint Logic Programming) and CHR (Constraint Handling Rules), both of which are available as SWI packages. Another approach is to limit the temporal scope of searches and prioritize recent frames in the search order over older frames.

Finally, and perhaps most significantly when considering online event monitoring, repeated queries that check for a certain event should only have to evaluate those frames which had not yet been processed by the same query earlier. In other words, if the Event Monitor has already determined that there is no match to a particular frame template for all frames up to 1000, and the video is currently at frame 1030, then VERSA should only have to check for a frame template match against frames 1001 through 1030.

### 6.3.2. Knowledge Base Size

Table 8 shows the Prolog used heap space after processing all frames of video. The used heap space is an approximate indication of the knowledge base size. Prior to asserting any facts, with only the VERSA and built-in SWI modules loaded, the used heap space is approximately 663,000 bytes (647K).

**Table 8: Knowledge Base Size (bytes)**

| Video Segment | Heap Space Used after Processing (bytes) |
|---|---|
| Left Bag | 2,996,000 |
| Left Box | 1,977,000 |
| Meet Crowd | 2,251,000 |
| Shopping Corridor 1 | 18,320,000 |
| Shopping Corridor 2 | 12,235,000 |
| Shopping Store Front | 4,274,000 |

### 6.3.3. Speedup From Cached Relationships

In Section 5.3.2., there is a discussion of caching the spatial relationships by asserting the relationships as facts into the Prolog knowledge base versus having to entail the spatial relationships for every query. This section presents an example of the query speedup on a representative query.

To compare, consider the following two queries that compute the same answer, where the first entails the spatial relationships on the fly and the second uses the cached relationships asserted into the knowledge base (differences shown in red text). The query is designed to detect two frames where a person has just crossed from being near a given object to no longer being near it. The queries were processed on the Left Bag video.

```
% Find an interval where a person has just crossed from being near
% a given object to no longer being near it.
% This version of the query does not use "cached relationships"
query1(F1,F2) :-
   match(F1,[object:O1,person:P1],[near(O1,P1)],[],1.0),
   match(F2, [object:O1, person:P1], [not_near(O1,P1)],[],1.0),
   F2 =< F1 + 5.

% This version of the query uses relationships previously asserted
% into the KB.
query1_kb(F1,F2) :-
   match(F1,[object:O1,person:P1],[near_kb(O1,P1)],[],1.0),
   match(F2, [object:O1, person:P1], [not_near_kb(O1,P1)],[],1.0),
   F2 =< F1 + 5.
```

The results of a **findall** query using these two versions to find all possible solutions are compared below. The built-in SWI-Prolog **time** predicate is used to compute the time to process each query.

```
?- time(findall( F1:F2, query1(F1,F2), Result1)).
> % 731,637 inferences, 5.54 CPU in 5.80 seconds (95% CPU, 132111
Lips)
> Result1 = [980:985, 981:985, 981:986, 982:985, 982:986, 982:987,
983:985, 983:986, ... : ...|...]

?- time(findall( F1:F2, query1_kb(F1,F2), Result1)).
```

```
> % 268,987 inferences, 0.95 CPU in 1.06 seconds (90% CPU, 282666 Lips)

> Result1 = [980:985, 981:985, 981:986, 982:985, 982:986, 982:987, 983:985, 983:986, ... : ...|...]
```

The first variant of the query requires 5.54 CPU seconds to process. The cached version requires only 0.95 CPU seconds to complete. The cached variant is nearly six times faster.

## 7. Conclusion

VERSA provides for a general purpose language and framework for defining events and recognizing them in live or recorded video streams. The approach for event recognition in VERSA is using a declarative logic language to define the spatial and temporal relationships that characterize a given event or activity. Doing so requires the definition of certain fundamental spatial and temporal relationships and a high-level syntax for specifying frame templates and query parameters. Although uncertainty handling in the current VERSA implementation is simplistic, the language and architecture is amenable to extending using Fuzzy Logic or similar approaches.

VERSA's high-level architecture is designed to work in XML-based, services-oriented environments. VERSA can be thought of as subscribing to the XML annotations streamed by a lower-level video analytics service that provides basic entity detection, labeling, and tracking using CVML or other XML-based video annotation standard. One or many VERSA Event Monitors could thus analyze video streams and provide alerts when certain events are detected.

## 7.1. Contribution

This thesis makes no claims on the relative performance or accuracy of event detection using VERSA in relation to any other method or implementation. VERSA provides for a general purpose language and framework for defining events and recognizing them in live or recorded video streams. Many other event detection methods use offline supervised training to recognize a small set of specific events. While it may be worthwhile to compare how a well-designed Event Template using VERSA compares to some of these special-purpose recognizers, that is not the main point of this work.

Rather the contribution of this work is in the flexibility of the logic-based method to represent a variety of events, the ability to generate Event Templates based on a Query-by-Example approach, and the application of certain structures, such as the *timestamp list* and Interval Sets to address certain temporal reasoning needs. Although at the highest level VERSA's architecture is not novel (see [47]), there is significant contribution in the details of the implementation, temporal reasoning capabilities, and the Query-by-Example prototype presented in this framework.

## 7.2. Limitations

There are some obvious limitations in this work as currently implemented. Firstly, spatial relationships are represented in two dimensions only. While sufficient for many uses, there are certain camera positions and perspectives that would be better served with some information regarding depth-of-field, which is lacking in VERSA currently. With some knowledge about the scene geometry, one could extend many of the 2D relationships into 3D equivalents.

As a rough approximation, one could even consider the relative scale differences in the bounding boxes as helping to indicate depth in the scene. In a typical surveillance video, should two people appear in the same scene, but the bounding box of one has 1/16 the area of the other, then one can reasonably infer that one person is further from the camera than the other. With some calibration, these scale differences could approximate depth, although clearly there would be confusion between very small people, such as toddlers, and scale differences due to distance from the camera.

VERSA's 2D spatial predicates are based only on rectangles and points. Arbitrary shaped polygons would help when trying to define areas of interest as static entities. For example, trapezoidal shapes can be helpful when attempting to define a flat area on the ground plane perspective.

VERSA does not currently provide mechanisms for dealing with tracks as first class objects in the system. The CVML annotation format provides no specification for handling track data, which must instead be entailed from the entity positions over time. Some events might best be expressed as logical reasoning about track relationships, and this would be an excellent future extension to VERSA. It would be very useful to be able to reason about the shape of tracks, intersections, etc.

The number of facts in the VERSA knowledge base grows linearly with the number of frames in the video and quadratically with the typical number of entities in each frame, if the number of binary intra-frame relationships is constant. After five minutes of video, even if limited to 10 frames per second, 3,000 frames will have been processed, generating hundreds-of-thousands of facts in the knowledge base. So queries become slower over time as more facts must be considered. Nothing is done in the current VERSA implementation to handle the ever-growing size of the knowledge base.

One immediate idea is to maintain a temporal window and discard old facts. Perhaps a better approach might be to reduce the sampling granularity of older material, scaling based on age. One keeps only a fraction of the frames in the KB where the fraction is a function of the age. A third approach would be to define certain interesting low-level events, such as the appearance of a given entity for the first time, and maintain a few frames around those foundational events indefinitely, but delete or scale back the “uninteresting” information based on a retention strategy. SWI-Prolog has interfaces to relational databases, so another strategy for maintaining query speed is to archive older facts to a database.

# Appendix A. CVML Data Sample

The following CVML data is from the first two frames of the “Left Bag 1” video of the CAVIAR data set.

```
<?xml version="1.0" encoding="UTF-8"?><dataset name="LeftBag">
   <frame number="0">
        <objectlist>
            <object id="0">
                <orientation>165</orientation>
                <box h="30" w="55" xc="184" yc="204"/>
                <appearance>appear</appearance>
                <hypothesislist>
                    <hypothesis evaluation="1.0" id="1" prev="0.0">
                        <movement evaluation="1.0">walking</movement>
                        <role evaluation="1.0">walker</role>
                        <context evaluation="1.0">immobile</context>
                        <situation evaluation="1.0">moving</situation>
                    </hypothesis>
                 </hypothesislist>
            </object>
            <object id="1">
                <orientation>147</orientation>
                <box h="18" w="26" xc="72" yc="76"/>
                <appearance>appear</appearance>
                <hypothesislist>
                    <hypothesis evaluation="1.0" id="1" prev="0.0">
                        <movement evaluation="1.0">walking</movement>
                        <role evaluation="1.0">walker</role>
                        <context evaluation="1.0">immobile</context>
                        <situation evaluation="1.0">moving</situation>
                    </hypothesis>
                 </hypothesislist>
            </object>
            <object id="2">
                <orientation>142</orientation>
                <box h="21" w="25" xc="78" yc="63"/>
                <appearance>appear</appearance>
                <hypothesislist>
                    <hypothesis evaluation="1.0" id="1" prev="0.0">
                        <movement evaluation="1.0">walking</movement>
                        <role evaluation="1.0">walker</role>
                        <context evaluation="1.0">immobile</context>
                        <situation evaluation="1.0">moving</situation>
                    </hypothesis>
                 </hypothesislist>
            </object>
        </objectlist>
        <grouplist>
            <group id="0">
                <orientation>59</orientation>
                <box h="32" w="32" xc="75" yc="69"/>
                <members>1,2</members>
                <appearance>appear</appearance>
                <hypothesislist>
                    <hypothesis evaluation="1.0" id="1" prev="0.0">
                        <movement evaluation="1.0">movement</movement>
                        <role evaluation="1.0">walkers</role>
                        <context evaluation="1.0">meeting</context>
```

```
                        <situation evaluation="1.0">moving</situation>
                    </hypothesis>
                 </hypothesislist>
            </group>
        </grouplist>
    </frame>
    <frame number="1">
        <objectlist>
            <object id="0">
                <orientation>165</orientation>
                <box h="27" w="55" xc="183" yc="200"/>
                <appearance>visible</appearance>
                <hypothesislist>
                    <hypothesis evaluation="1.0" id="1" prev="1.0">
                        <movement evaluation="1.0">walking</movement>
                        <role evaluation="1.0">walker</role>
                        <context evaluation="1.0">immobile</context>
                        <situation evaluation="1.0">moving</situation>
                    </hypothesis>
                 </hypothesislist>
            </object>
            <object id="1">
                <orientation>147</orientation>
                <box h="19" w="25" xc="71" yc="76"/>
                <appearance>visible</appearance>
                <hypothesislist>
                    <hypothesis evaluation="1.0" id="1" prev="1.0">
                        <movement evaluation="1.0">walking</movement>
                        <role evaluation="1.0">walker</role>
                        <context evaluation="1.0">immobile</context>
                        <situation evaluation="1.0">moving</situation>
                    </hypothesis>
                 </hypothesislist>
            </object>
            <object id="2">
                <orientation>142</orientation>
                <box h="21" w="25" xc="78" yc="63"/>
                <appearance>visible</appearance>
                <hypothesislist>
                    <hypothesis evaluation="1.0" id="1" prev="1.0">
                        <movement evaluation="1.0">walking</movement>
                        <role evaluation="1.0">walker</role>
                        <context evaluation="1.0">immobile</context>
                        <situation evaluation="1.0">moving</situation>
                    </hypothesis>
                 </hypothesislist>
            </object>
        </objectlist>
        <grouplist>
            <group id="0">
                <orientation>65</orientation>
                <box h="33" w="32" xc="75" yc="69"/>
                <members>1,2</members>
                <appearance>visible</appearance>
                <hypothesislist>
                    <hypothesis evaluation="1.0" id="1" prev="1.0">
                        <movement evaluation="1.0">movement</movement>
                        <role evaluation="1.0">walkers</role>
                        <context evaluation="1.0">meeting</context>
                        <situation evaluation="1.0">moving</situation>
                    </hypothesis>
                 </hypothesislist>
            </group>
```

```
        </grouplist>
    </frame>
</dataset>
```